\documentclass[10pt]{article} 
\usepackage[accepted]{rlc}
\setcitestyle{authoryear,round,citesep={,},aysep={},yysep={;}}

\usepackage{amssymb}            
\usepackage{mathtools}          
\usepackage{mathrsfs}           
\mathtoolsset{showonlyrefs}     
\usepackage{graphicx}           
\usepackage{subcaption}         
\usepackage[space]{grffile}     
\usepackage{url}                

\usepackage{wrapfig}    
\usepackage{adjustbox}

\title{
Revisiting Sparse Rewards for \\ Goal-Reaching Reinforcement Learning
}

\author{$\text{Gautham Vasan}^{\dagger \ddagger}$ \\
    vasan@ualberta.ca \\
    \And
    $\text{Yan Wang}^{\dagger \ddagger}$  \\
    yan28@ualberta.ca \\
    \And
    $\text{Fahim Shahriar}^{\dagger \ddagger}$ \\
    fshahri1@ualberta.ca \hspace{0.2cm} \\
    \And
    $\text{James Bergstra}^\zeta$  \\
    james.bergstra@ocado.com \\
    \And
    $\text{Martin Jagersand}^{\ddagger}$  \\
    jag@ualberta.ca \\
    \And
    $\text{A. Rupam Mahmood}^{\iota \dagger \ddagger}$  \\
    armahmood@ualberta.ca \\
    \And
    \\
    $^{\dagger}$ Alberta Machine Intelligence Institute (Amii), Edmonton, Canada \\    
    $^{\ddagger}$ Department of Computing Science, University of Alberta, Edmonton, Canada \\
    $^{\iota}$ Canada CIFAR AI Chair \\
    $^{\zeta}$ AI Platform, Ocado Technology, Toronto, Canada 
}

\begin{document}

\maketitle

\begin{abstract}
Many real-world robot learning problems, such as pick-and-place or arriving at a destination, can be seen as a problem of reaching a goal state as soon as possible. 
These problems, when formulated as episodic reinforcement learning tasks, can easily be specified to align well with our intended goal:  $-1$ reward every time step with termination upon reaching the goal state, called \emph{minimum-time} tasks. 
Despite this simplicity, such formulations are often overlooked in favor of dense rewards due to their perceived difficulty and lack of informativeness. 
Our studies contrast the two reward paradigms, revealing that the minimum-time task specification not only facilitates learning higher-quality policies but can also surpass dense-reward-based policies on their own performance metrics. 
Crucially, we also identify the goal-hit rate of the initial policy as a robust early indicator for learning success in such sparse feedback settings. 
Finally, using four distinct real-robotic platforms, we show that it is possible to learn pixel-based policies from scratch within two to three hours using constant negative rewards.
Our video demo can be found \href{https://youtu.be/a6zlVUuKzBc}{here}.\footnote{Video demo: \url{https://youtu.be/a6zlVUuKzBc} \\ Code for simulation experiments: \url{https://github.com/gauthamvasan/rl_suite} \\
Code for robot experiments: \url{https://github.com/rlai-lab/relod} }
\end{abstract}

\section{Introduction}
\label{sec:introduction}

\begin{figure}[h]
    \centering
    \vspace{-10pt}
    \begin{subfigure}{0.48\columnwidth}
        \includegraphics[width=\columnwidth]{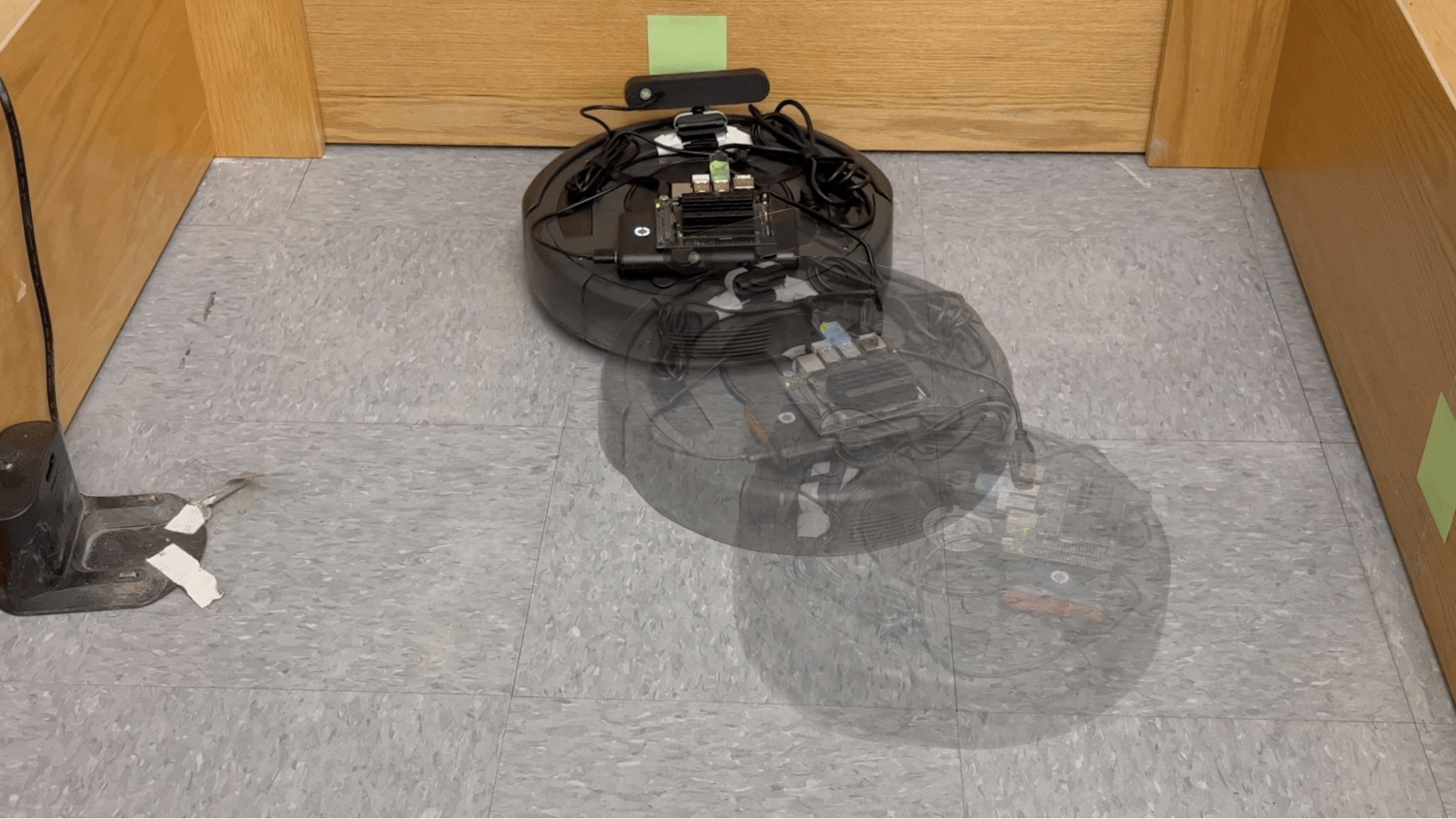}
        \caption{Create-Reacher}
        \label{fig:create_overlay}
    \end{subfigure}
    \begin{subfigure}{0.48\columnwidth}
        \includegraphics[width=\columnwidth]{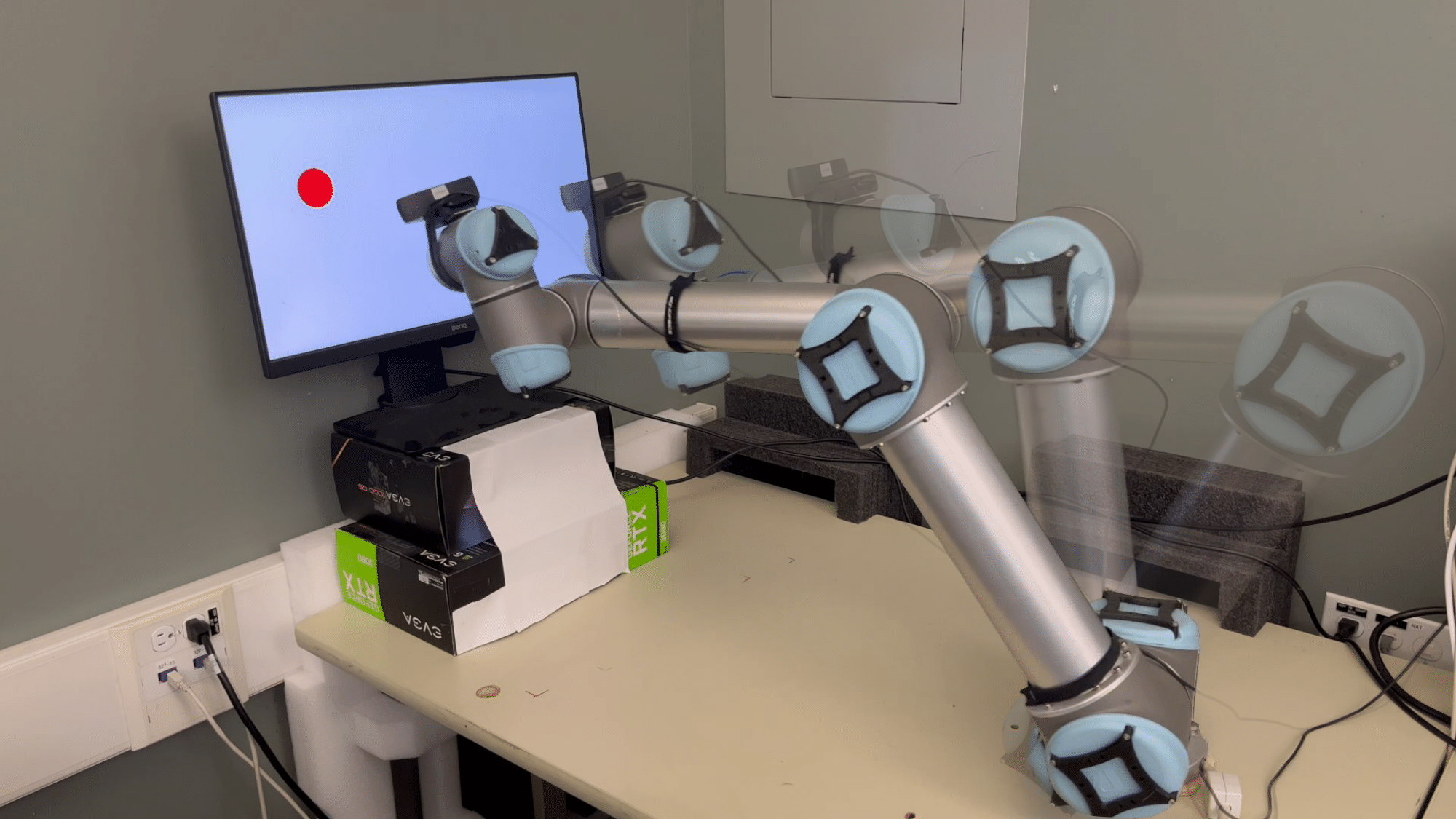}
        \caption{UR5-VisualReacher}
        \label{fig:ur5_overlay}
    \end{subfigure}
    \begin{subfigure}{0.48\columnwidth}
        \includegraphics[width=\columnwidth]{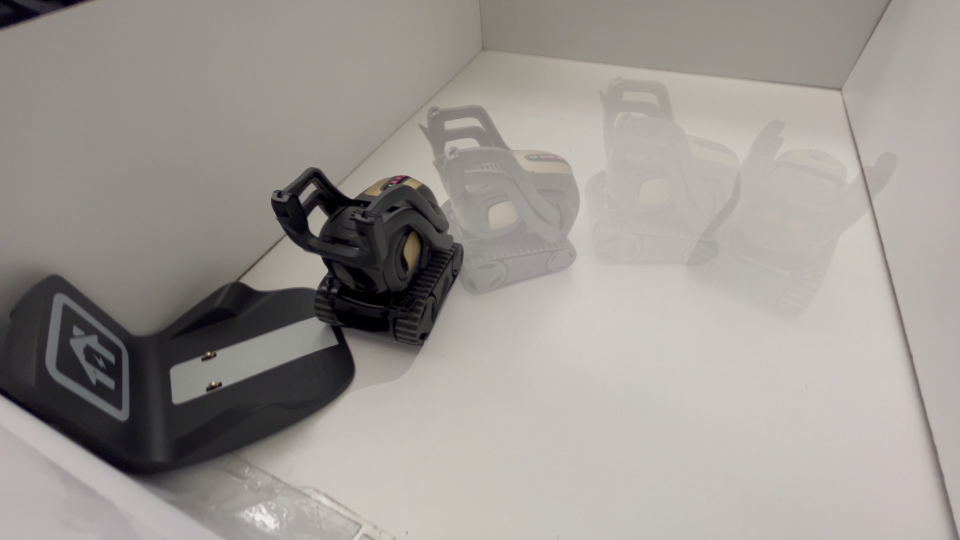}
        \caption{Vector-ChargerDetector}
        \label{fig:vector_overlay}
    \end{subfigure}
    \begin{subfigure}{0.48\columnwidth}
        \includegraphics[width=\columnwidth]{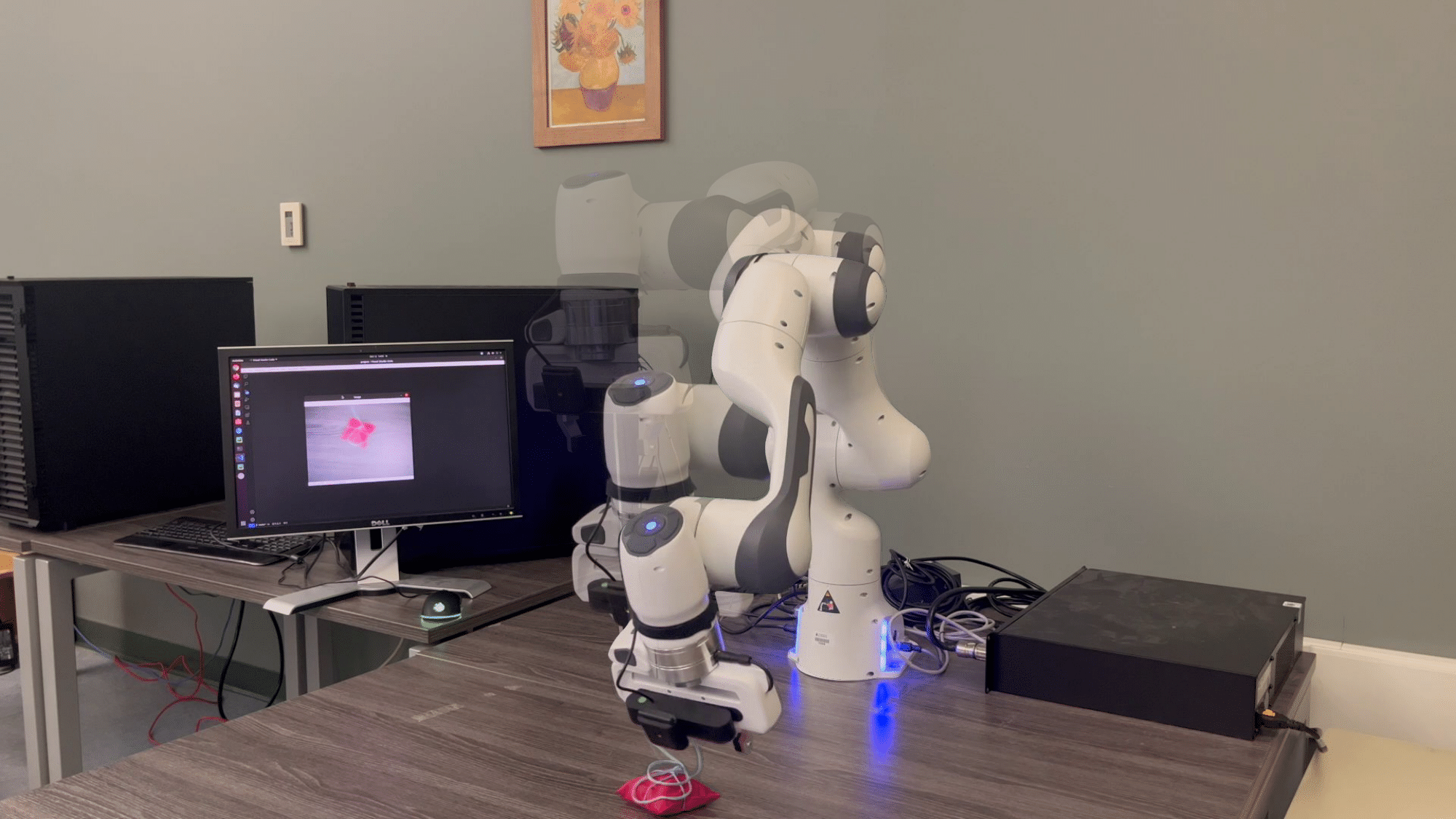}
        \caption{Franka-VisualReacher}
        \label{fig:franka_overlay}
    \end{subfigure}
    \caption{Real World Robot Tasks}
    \label{fig:overlay}
    \vspace{-10pt}
\end{figure}

In Reinforcement Learning (RL), the task designer implicitly specifies the desired behavior using the reward signal.
In order to guide the learning agent to a reasonable solution, task designers often rely on task-specific domain knowledge to hand-craft a dense reward function with state-to-state differences that facilitate faster learning.
Since guiding rewards reflect the task designer's preferred behaviors, they could bias the solutions that the agent finds, potentially leading to sub-optimal outcomes \citep{riedmiller2018learning}.
For most goal-reaching problems, there is a simpler alternative that is easy to specify but still incorporates our intended goal accurately: a constant negative reward every time step, with termination upon reaching the goal state.
We call this the \emph{minimum-time} specification since maximizing the undiscounted sum of these rewards leads to reaching the goal state as soon as possible.
This specification avoids biasing the final solution, focusing solely on recognizing task success.
This approach is also suitable for specifying unforeseen goal-reaching tasks where a human instructor may be unable to provide domain knowledge beforehand.

An alternative and simple specification of rewards is to give a discounted positive reward upon ``success'' and zero everywhere else.
Such a reward is often considered a \emph{sparse reward}, as a differentiating reward signal is available only sparsely.
The minimum-time reward can be seen as an extreme form of sparse reward, where the reward value never changes and the differentiating signal only comes from episodic accumulated rewards, called the \emph{return}.
However, minimum-time specification has the advantage of being simpler as it does not require discounting as part of the problem.

Despite its simplicity and ease of specification, minimum-time tasks are generally thought to be hard to solve \citep{andrychowicz2017hindsight}. 
Guiding reward formulations offer state-by-state reward differences, which can be informative for quick policy improvement. 
This can provide early signs of learning, allowing us to determine whether learning can occur quickly. 
Sparse-reward tasks may take much longer to show any signs of learning, as informative signals are given only sparsely; a robot may only rarely stumble upon the goal via trial-and-error interactions  \citep{kober2013reinforcement}.

In this paper, we conduct a series of carefully designed studies to compare dense-reward and minimum-time formulations in RL. 
Our research confirms the popular notion that dense-reward formulations facilitate faster learning in comparison to sparse-reward formulations. 
However, our findings also demonstrate that agents trained with minimum-time specification outperform those trained with dense rewards, even in terms of dense-reward performance metrics. 
In particular, we establish the superiority of the minimum-time formulation over the guiding reward formulation on tasks involving reaching with simulated and industrial robotic arms, both in terms of ease of specification and its ability to attain the desired final behavior. 
We identify that RL agents can quickly and reliably solve complex vision-based tasks in the minimum-time formulation \emph{if} the agent can reach the goal often enough using its initial policy. 
Leveraging our insights, we successfully set up vision-based reaching tasks with sparse rewards on four distinct real robots and demonstrate their ability to learn a pixel-based control policy from scratch within two to three hours (see Fig.\ \ref{fig:overlay}).

\section{Related Work}
\label{sec:related}
\vspace{-10pt}
\textbf{Reward shaping} Learning with sparse rewards can be challenging since the agent has to explore the environment extensively to discover the right set of actions that lead to the reward.
Hence, task designers often manually craft a reward function through trial-and-error such that it maximizes both the task performance metric and enables fast learning \citep{vasan2017learning, mahmood2018benchmarking, lee2019making, knox2023reward}.
\cite{ma2023eureka} proposed using a large language model to automate reward design in various tasks using task-specific prompts, reward templates and evolutionary search.
Recent work by \cite{rocamonde2023vision} and \cite{wang2024rl} explores using large pre-trained vision-language models (VLMs) to design reward functions for RL tasks. These approaches, akin to neural architecture search, incur high computational costs.
VLMs face challenges like sensitivity to visual realism, hallucinations, scalability issues, and mis-alignments between visual and language modalities.
They often struggle to provide precise, task-relevant feedback necessary for effective RL, especially in tasks requiring detailed spatial understanding or abstract reasoning.

Ad hoc reward design often leads to \emph{reward shaping}, where the reward function is used to communicate the underlying performance metric \textit{and} to steer the agent's learning towards a desired policy \citep{ng1999policy}.
For example, \cite{mataric1994reward} proposed rewarding the agent for taking steps up the gradient rather than just for achieving the final goal to speed up learning.
However, it is also widely recognized that ad hoc reward shaping is unsafe as it may alter the optimal solution for a particular RL task \citep{amodei2016concrete, knox2023reward}.

Although guiding reward design seems like a sound approach, \cite{booth2023perils} show that relying on trial-and-error to design guiding rewards can result in over-fitting, where reward functions are inadvertently excessively tailored for use with a specific algorithm.
Rewards should only communicate to the agent \textit{what} we want achieved rather than detailing \textit{how} we want it achieved \citep{sutton2018reinforcement}. In this work, we adhere to this principle and sidestep the complexities of reward function (mis)-design, thus focusing solely on making learning with sparse rewards tractable.

\textbf{Learning from sparse rewards} \cite{riedmiller2018learning} introduced SAC-X, a method for learning from sparse rewards, which learns policies for auxiliary tasks concurrently with the main task to explore the observation space efficiently. \cite{hertweck2020simple} extended this by proposing agent-internal auxiliary tasks to enhance exploration in sparse reward settings, particularly for tasks like Ball-in-Cup, using only raw sensor data. \cite{andrychowicz2017hindsight} presented Hindsight Experience Replay (HER) for learning from failed episodes by treating seen states as pseudo-goals, applicable to environments with multiple goal states but limited in vision-based tasks. \cite{nair2018visual} expanded HER for vision-based tasks like Reaching and Pushing but with constraints on camera setup and applicability to 2D plane environments.
\cite{korenkevych2019autoregressive} use autoregressive processes for smoother exploration compared to Gaussian policies.
While these approaches use novel strategies to improve exploration in sparse reward scenarios, our work differs by predicting whether a minimum-time task can be effectively learned from scratch based on initial policy performance.

\textbf{Minimum time problems} in optimal control aim to transfer a system from an initial to a final state in the shortest time, with the cost function being the time taken between steps \citep[Chapter~9]{chui2012linear}.
For example, \cite{penicka2022learning} combines classical path planning with model-free deep RL to optimize a neural network policy for minimum-time flight of a quadrotor through a sequence of waypoints with obstacles.
Control benchmarks in RL like MountainCar \citep{sutton2018reinforcement} and DotReacher \citep{garg2021alternate} also use the minimum-time formulation with a reward of $-1$ each step until termination.

\section{How to Specify Goal-Reaching Tasks in RL?}
\label{sec:specification}
\begin{wrapfigure}{r}{.3\columnwidth}
    \centering
    \includegraphics[width=.3\columnwidth]{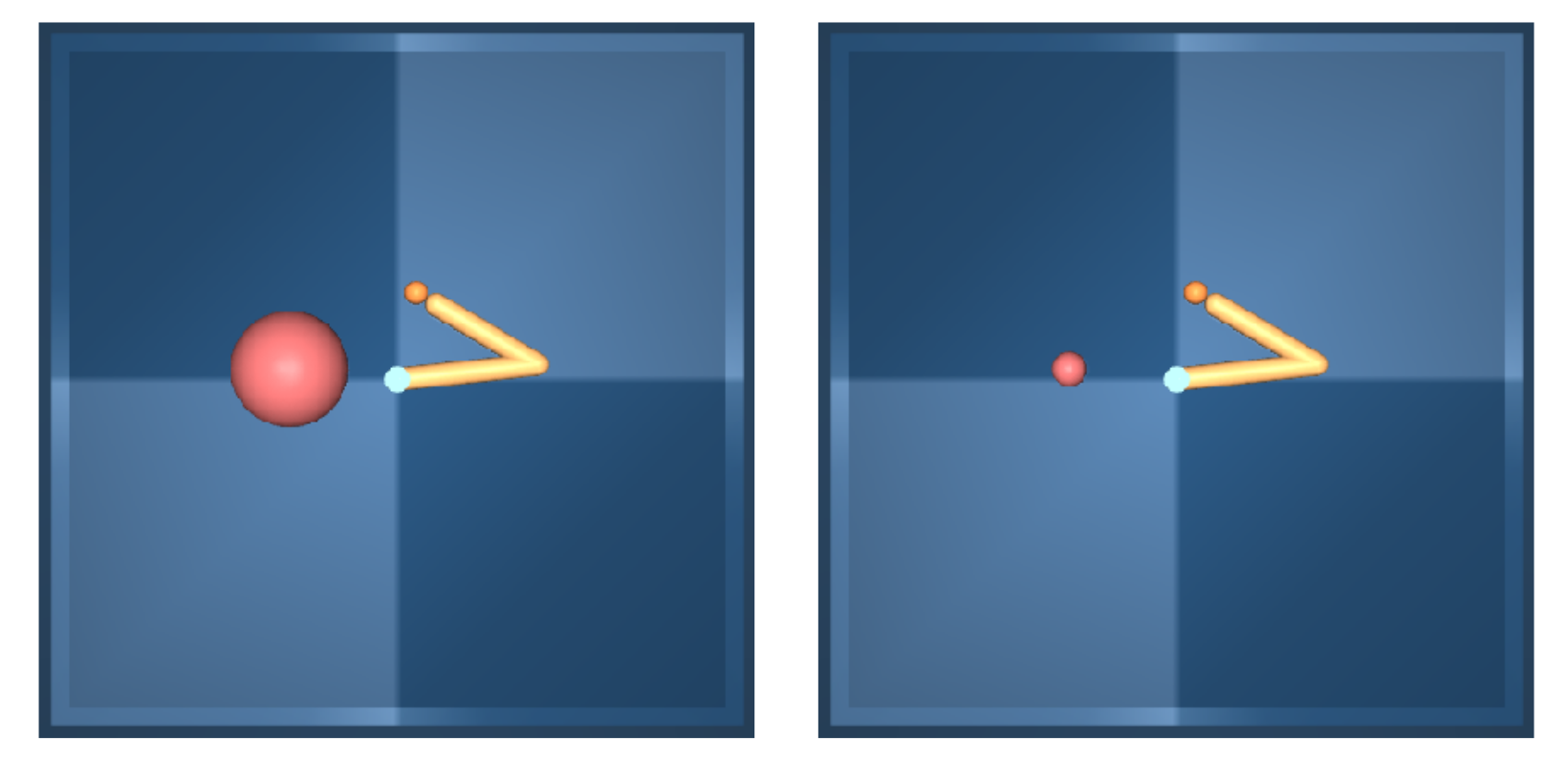}
    \caption{Reacher-Easy (left) and Reacher-Hard (right) from DeepMind Control Suite }
    \label{fig:reacher_task}
\end{wrapfigure}

In this section, we explore various formulations of goal-reaching tasks as discussed in the literature.
We use the classic Reacher problem as an illustrative example (Fig. \ref{fig:reacher_task}). 
The reaching task using a two-link robot arm aims to move the fingertip of a planar arm with two degrees of freedom to a random spherical target on a 2D plane  \citep{tassa2018deepmind}. 
In this scenario, \emph{we hold an abstract concept wherein the objective is for the robot arm to swiftly reach a designated target state and remain there}.
The difficulty of the Reacher task is conditional on the size of the target.

\textbf{Background} A Markov Decision Process (MDP) is used to model the agent-environment interaction, where an agent interacts with its environment at discrete timesteps.
The agent takes an action $A_t \in \mathcal{A}$ at state $S_t \in \mathcal{S}$ every timestep ${t}$ using a probability distribution $\pi$ called a \emph{policy}: $A_t\sim\pi(\cdot | S_t)$. 
After executing the action, the agent receives the next state $S_{t+1}$ at the subsequent timestep ${t+1}$ and a reward $R_{t+1}$ according to a transition probability density function $S_{t+1}, R_{t+1} \sim p(\cdot, \cdot | S_t, A_t)$. 
We use continuous state space $\mathcal{S}$ and action space $\mathcal{A}$ for all our tasks.

The \emph{observation space} consists of the position and velocity of the fingertip, and the vector from the fingertip to the target. The \emph{action space} is the torques to be applied to the two joints.
We now introduce three specifications based on the choice of reward function and termination.

\begin{table}[htbp]
    \centering
    \begin{adjustbox}{width=0.99\textwidth}
        \begin{tabular}{|c|c|c|}
        \hline
        \textbf{Guiding Reward Formulation} & \textbf{Contact Reward Formulation} & \textbf{Minimum-Time Formulation} \\
        \hline
       $R_t = \begin{cases}
            1 &\text{if in target},\\
            -||x_{goal} - x_{pos}|| &\text{otherwise}.
            \end{cases}
        $ &
        $R_t = \begin{cases}
        1 &\text{if in target},\\
        0 &\text{otherwise}.
        \end{cases}$ &
        $R_t = -1 \text{\; (until reaching the target)}$ \\
        \hline
        Fixed length episodes ($T=1000$) &  Fixed length episodes ($T=1000$) & 
        \begin{tabular}[x]{@{}c@{}} Varying length episodes; Termination upon\\reaching  the goal with near zero velocity\end{tabular} \\
        \hline
        \end{tabular}
    \end{adjustbox}
    \caption{Task Formulations: Choices for reward function and termination conditions}
    \label{tab:reward}
    \vspace{-7pt}
\end{table}

\textbf{Guiding reward formulation}
The reward function is inspired by \cite{brockman2016openai}, where reacher is incentivized to get as close to the target as possible. 
We also include a commonly used precision reward term previously used on a real robot reacher task using the UR5 industrial arm \citep{lan2022model, farrahi2023reducing, che2023correcting, elsayed2024revisiting}. 
The reward function has two components: (i) penalty term calculated as the negative of the Euclidean distance between the fingertip and the target, and (ii) precision term where the agent receives $+1$ reward when the fingertip is inside the target (see Table \ref{tab:reward}). Each episode has a time limit of 1000.
The time limit is set long enough to incentivize the agent to remain at the goal once reached.

\textbf{Contact Reward Formulation}
The agent will receive a reward of $+1$ whenever it is within the target sphere and $0$ otherwise (see Table \ref{tab:reward}).
Each episode has a time limit of $1000$.
This choice of reward and timeout follows the \textit{dm\_control} suite \citep{tassa2018deepmind}.

\textbf{Minimum-Time formulation}
Shorter episodes are encouraged by imposing a reward of $-1$ for each step until termination. 
The episode terminates when the fingertip of the arm reaches the target with near-zero velocity.
An episode is completed only when the agent reaches the goal state. 
If the agent exceeds the time limit, we reset the environment such that the goal state remains the same, but the robot arm moves to a different starting state. 
Note that, unlike the previous formulations, we get rid of the fixed time limit horizon, where all episodes have a uniform episode length. 

Since resetting the agent has an associated time cost in the real world, we also penalize the agent when it fails to reach the goal state within the time limit (e.g., $-K$ where $K$ timesteps is required to reset the agent). 
The episodic returns and lengths are adjusted appropriately. 
For example, consider a task where the reward is $-1$ each timestep, the time limit is $100$ steps, and $\text{reset penalty}=-20$.
If an agent times out thrice consecutively and finally reaches the goal in $25$ steps since the last timeout, then the return of the episode, in this case, is $-100  - 20  - 100  - 20  - 100  - 20  - 25 = -385$, and the length of the episode is $385$.
Our learning curves use undiscounted returns estimated in this manner.

\section{Which Formulation Achieves the Desired Reaching Behavior?}
\begin{figure}[h]
\centering
    \begin{subfigure}{.32\textwidth}
        \includegraphics[width=\columnwidth]{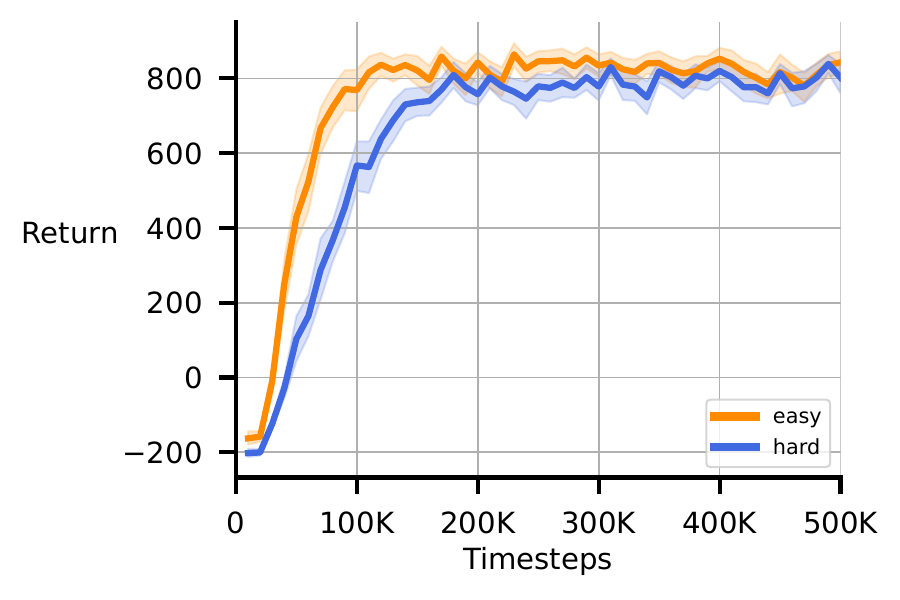}
        \caption{Guiding reward formulation}
        \label{fig:ar_reacher}
    \end{subfigure}
    \begin{subfigure}{.32\textwidth}
        \includegraphics[width=\columnwidth]{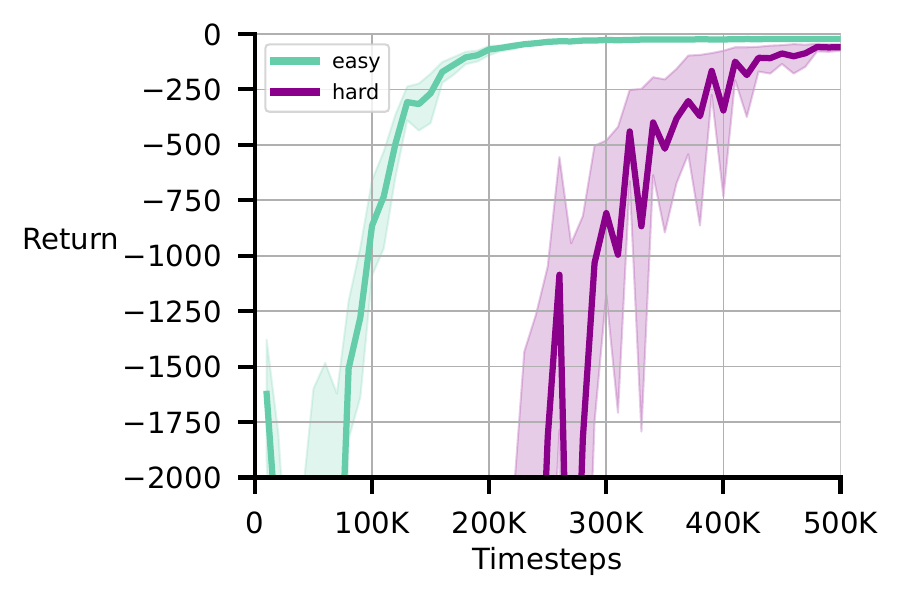}
        \caption{Minimum-time formulation}
        \label{fig:vt_reacher}
    \end{subfigure}
    \begin{subfigure}{.32\textwidth}
        \includegraphics[width=\columnwidth]{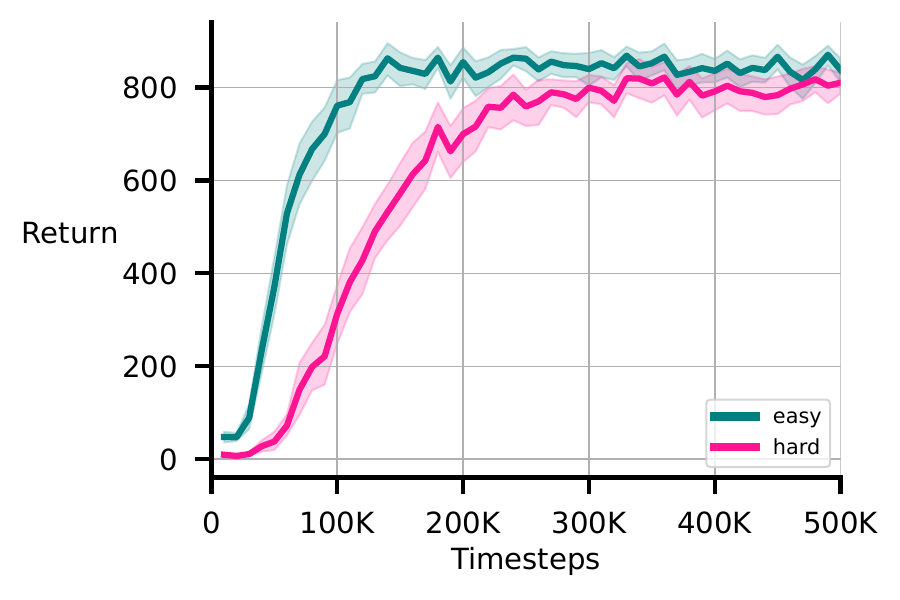}
        \caption{Contact reward formulation}
        \label{fig:ftl_reacher}
    \end{subfigure}
    \caption{Each solid learning curve is an average of 30 independent runs. The shaded regions represent a 95\% confidence interval}
    \label{fig:reacher}
    \vspace{-10pt}
\end{figure}

We use a state-of-the-art RL method---
Soft Actor-Critic (SAC) to to solve each formulation of the Reacher task \citep{haarnoja2018soft}. 
We use the same hyper-parameters (listed in \ref{sec:appendix_A}) to train all variants.
From Fig. \ref{fig:ar_reacher}, it is clear that both Reacher-Easy and Reacher-Hard agents learn fast with the guiding reward formulation. 
While the speed of learning is not as rapid, it is still fast with the contact reward formulation (see Fig. \ref{fig:ftl_reacher}).
In contrast, when utilizing the minimum-time formulation (see Figure \ref{fig:vt_reacher}), the learning curves exhibit a slower progression and more pronounced variability.

\begin{wrapfigure}{r}{.5\textwidth}
    \centering
    \begin{adjustbox}{width=0.5\textwidth}
    \begin{tabular}{|c|c|c|c|}
        \hline
         & \begin{tabular}{@{}c@{}} \textbf{Guiding Reward} \\ \textbf{Formulation}\end{tabular} & \begin{tabular}{@{}c@{}} \textbf{Contact Reward} \\ \textbf{Formulation}\end{tabular} &\begin{tabular}{@{}c@{}} \textbf{Minimum-Time} \\ \textbf{Formulation}\end{tabular} \\ \hline
        \begin{tabular}{@{}c@{}} Ease of \\ specification \end{tabular} & $\times$ & $\checkmark$ & $\checkmark$ \\ \hline
        \begin{tabular}{@{}c@{}} Fast \\ Learning \end{tabular} & $\checkmark$ & $\sim$ & $\times$ \\ \hline
        \begin{tabular}{@{}c@{}} Superior Final\\Performance \end{tabular} & $\times$ & $\times$ & $\checkmark$  \\
        \hline
    \end{tabular}
    \end{adjustbox}
    \caption{Evaluating each task specification}
    \label{tab:metrics}
    \vspace{-10pt}
\end{wrapfigure}
\paragraph{Evaluating the final learned behaviours}
\label{sec:eval_final}
For a fair comparison of the three task specifications, we evaluate the final learned policies using three metrics (see Table. \ref{tab:metrics}):
1) \emph{Ease of specification}: Do you require any domain knowledge to craft a reward function?
2) \emph{Fast learning}: Is learning quick and robust during training?
3) \emph{Superior Final Performance}: Once the training phase concludes, which version demonstrates the most desirable final learned behavior, as intended by the task designer? 

To answer which formulation achieves superior final performance, we measure: 1) the number of time steps needed to reach the target (termed “steps \textit{to} goal”), and 2) the duration the agent remains at the target once reached (termed “steps \textit{on} goal”).
We evaluate all $30$ learned policies from each formulation ($90$ runs in total) on $500$ episodes, lasting $5K$ timesteps each ($2.5M$ steps in total).

\begin{figure}[h]
    \centering
    \begin{subfigure}{.24\textwidth}
        \includegraphics[width=\columnwidth]{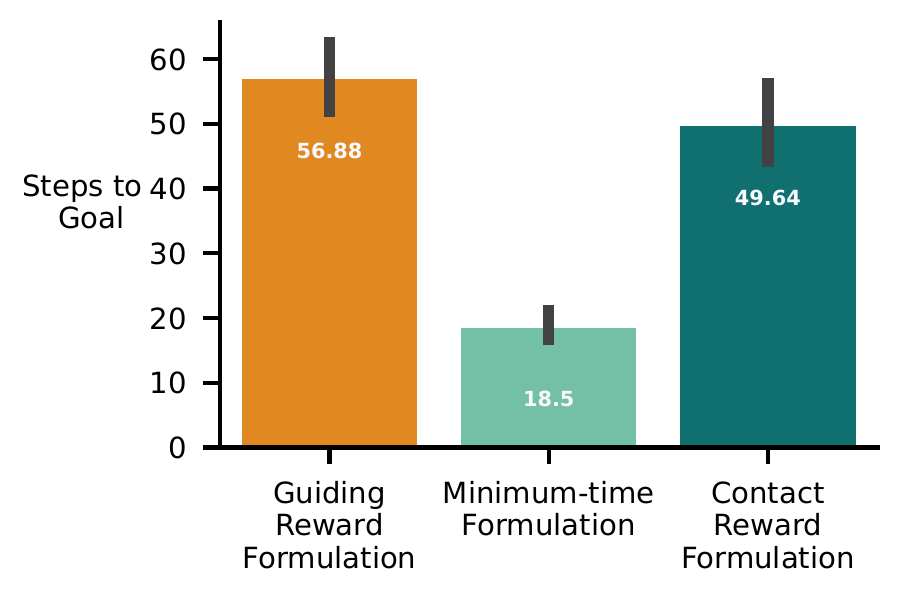}
        \caption{Reacher-Easy}
        \label{fig:reacher_easy_steps_to_goal}
    \end{subfigure}
    \begin{subfigure}{.24\textwidth}
        \includegraphics[width=\columnwidth]{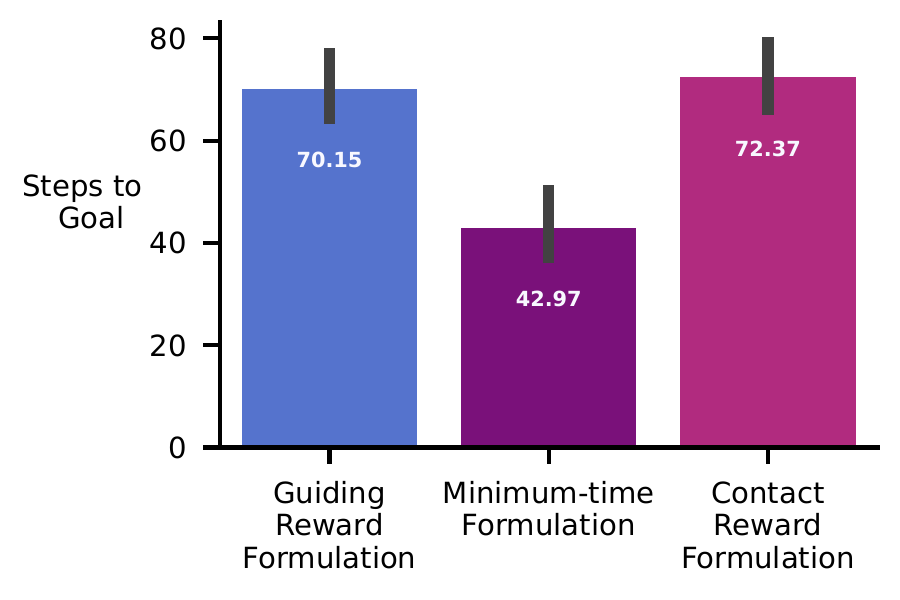}
        \caption{Reacher-Hard}
        \label{fig:reacher_hard_steps_to_goal}
    \end{subfigure}
    \begin{subfigure}{.24\textwidth}
        \includegraphics[width=\columnwidth]{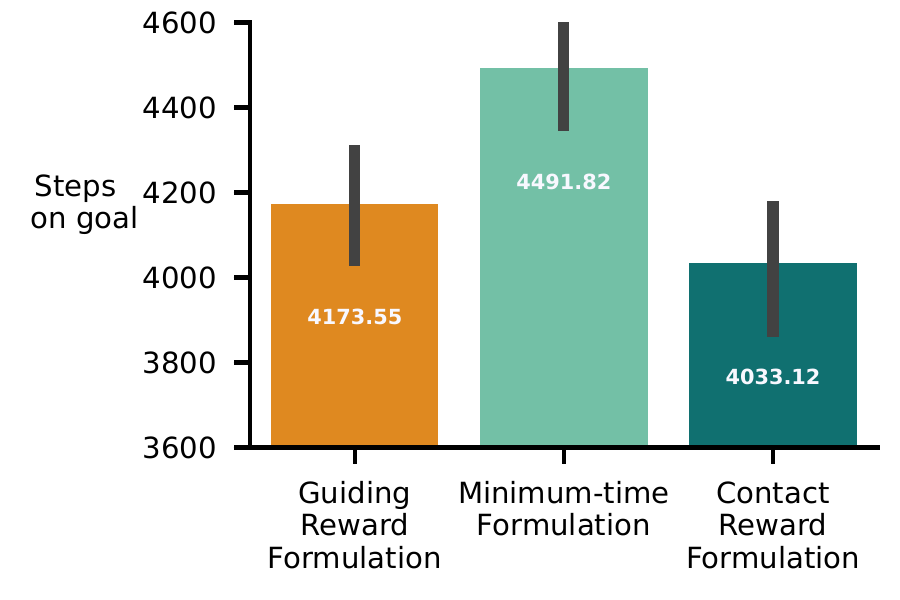}
        \caption{Reacher-Easy}
        \label{fig:reacher_easy_steps_on_goal}
    \end{subfigure}
    \begin{subfigure}{.24\textwidth}
        \includegraphics[width=\columnwidth]{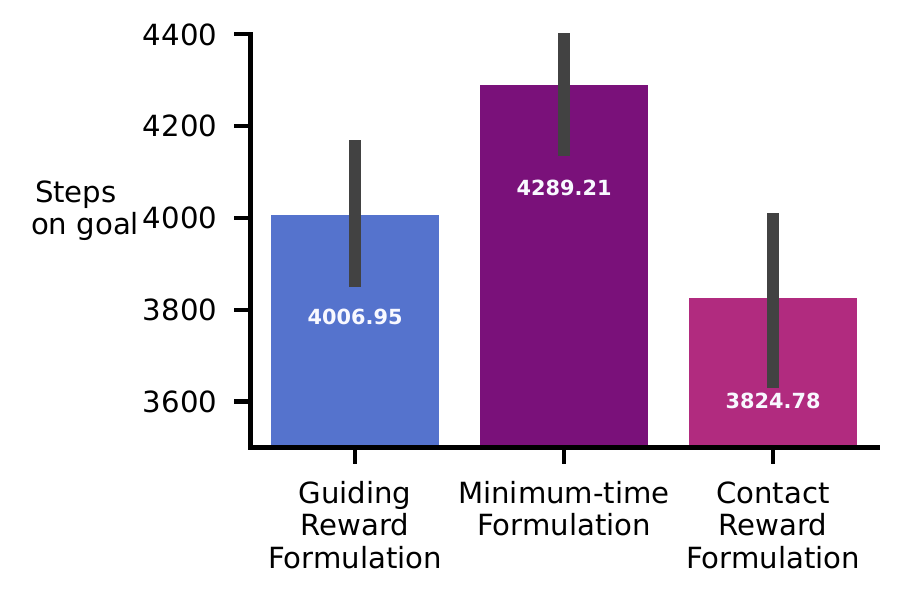}
        \caption{Reacher-Hard}
        \label{fig:reacher_hard_steps_on_goal}
    \end{subfigure}
    \caption{Comparison of the three formulations of the reaching task using the number of steps it requires to reach the goal (\ref{fig:reacher_easy_steps_to_goal}, \ref{fig:reacher_hard_steps_to_goal}) and the number of steps it stays within the target sphere upon reaching it (\ref{fig:reacher_easy_steps_on_goal}, \ref{fig:reacher_hard_steps_on_goal}). The standard error is estimated over 30 independent runs. In bar charts \ref{fig:reacher_easy_steps_to_goal} \& \ref{fig:reacher_hard_steps_to_goal}, a lower value signifies better performance. In \ref{fig:reacher_easy_steps_on_goal} \& \ref{fig:reacher_hard_steps_on_goal}, a higher value signifies better performance. }
    \label{fig:steps_to_goal}
    \vspace{-5pt}
\end{figure}

While the guiding reward formulation achieves faster learning compared to guiding reward and minimum-time formulation (Fig. \ref{fig:reacher}), it is not easy to specify.
Despite slower learning and less visually appealing learning curves, the final learned performance shows surprisingly superior quality and robustness when using the minimum-time formulation (Fig. \ref{fig:steps_to_goal}): $\sim 3X$ faster on Reacher-Easy and $\sim 2X$ faster on Reacher-Hard (Fig. \ref{fig:reacher_easy_steps_to_goal} \& \ref{fig:reacher_hard_steps_to_goal}). 
The minimum-time policy also stays inside the target area longer, resulting in over 500 higher accrued rewards (Fig. \ref{fig:reacher_easy_steps_on_goal} \& \ref{fig:reacher_hard_steps_on_goal}). 

\begin{figure}[htp]
    \centering
    \begin{subfigure}{.32\textwidth}
        \includegraphics[width=\columnwidth]{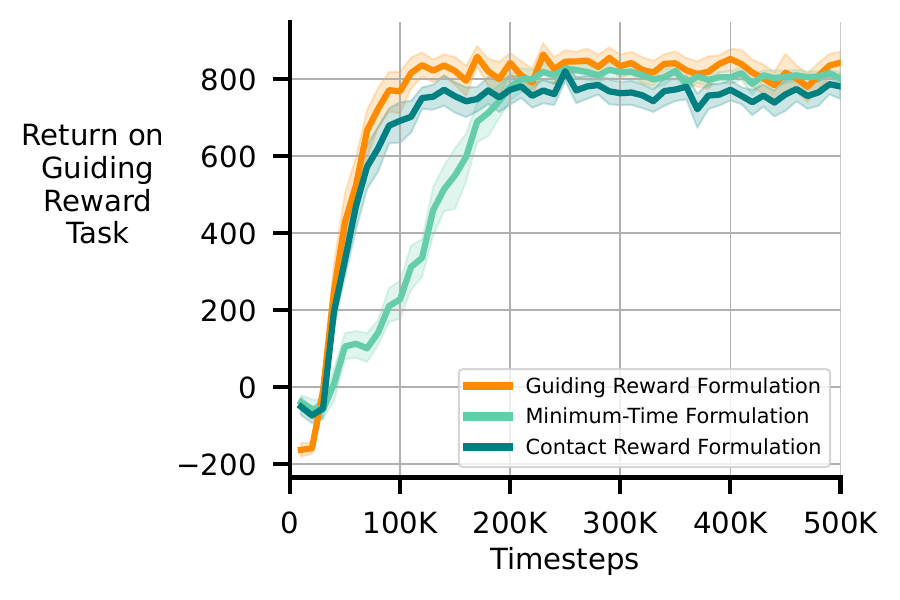}
        \caption{Guiding Reward task}
        \label{fig:ar_easy_eval_across_tasks}
    \end{subfigure}
    \begin{subfigure}{.32\textwidth}
        \includegraphics[width=\columnwidth]{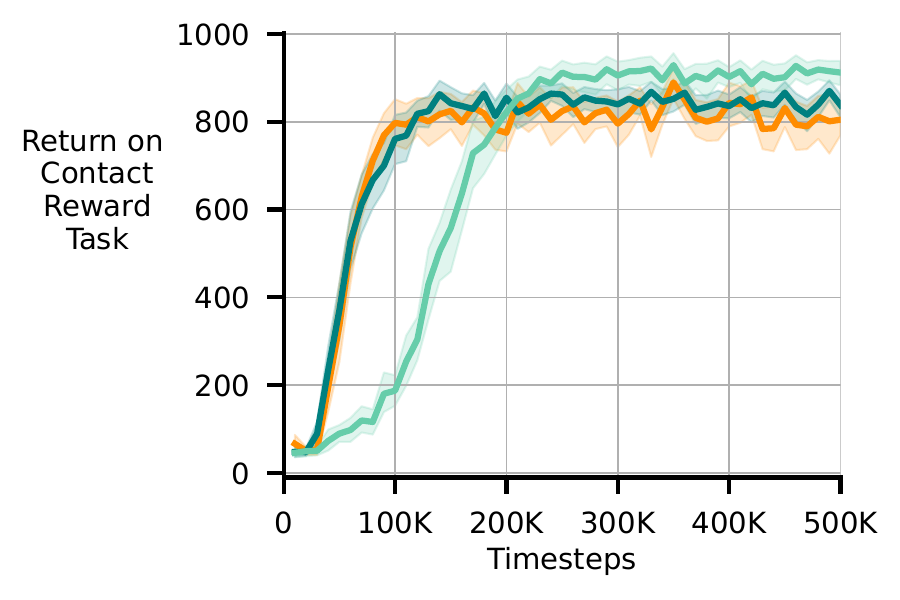}
        \caption{Contact Reward task}
        \label{fig:ftl_easy_eval_across_tasks}
    \end{subfigure}
    \begin{subfigure}{.32\textwidth}
        \includegraphics[width=\columnwidth]{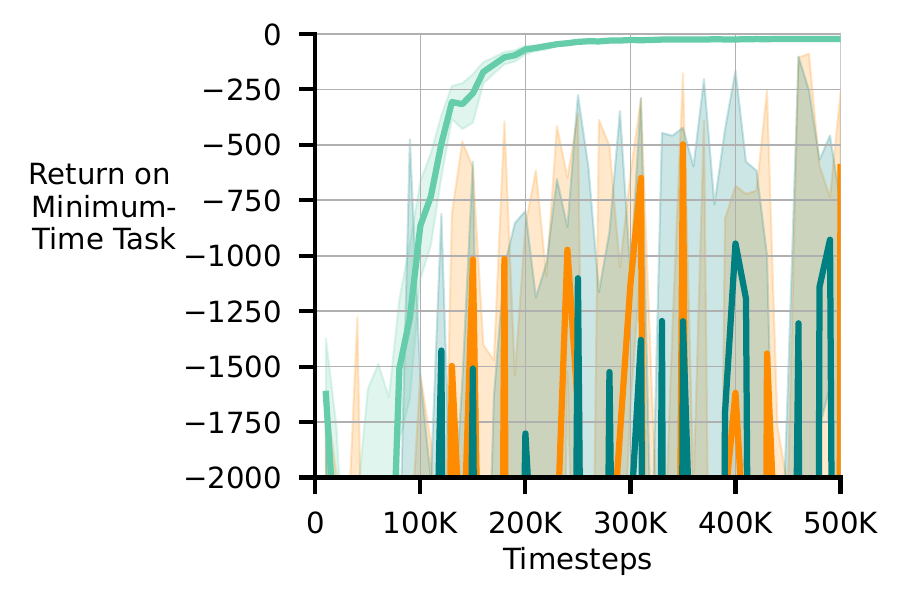}
        \caption{Minimum-Time task}
        \label{fig:vt_easy_eval_across_tasks}
    \end{subfigure}
    \newline
    \begin{subfigure}{.32\textwidth}
        \includegraphics[width=\columnwidth]{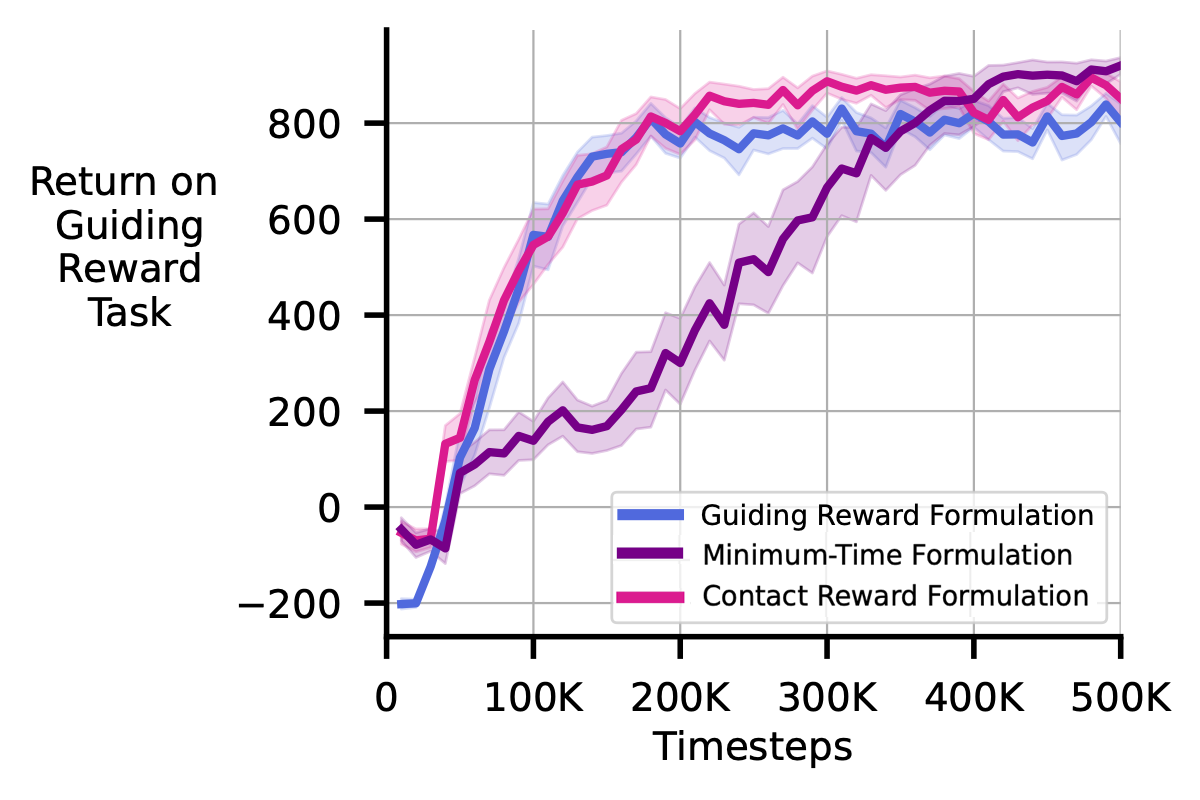}
        \caption{Guiding Reward task}
        \label{fig:ar_hard_eval_across_tasks}
    \end{subfigure}
    \begin{subfigure}{.32\textwidth}
        \includegraphics[width=\columnwidth]{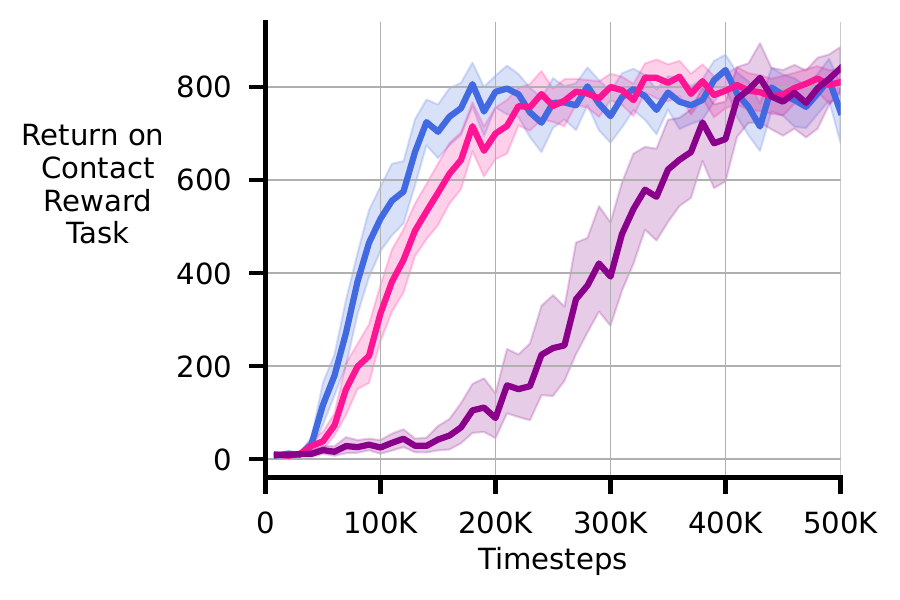}
        \caption{Contact Reward task}
        \label{fig:ftl_hard_eval_across_tasks}
    \end{subfigure}
    \begin{subfigure}{.32\textwidth}
        \includegraphics[width=\columnwidth]{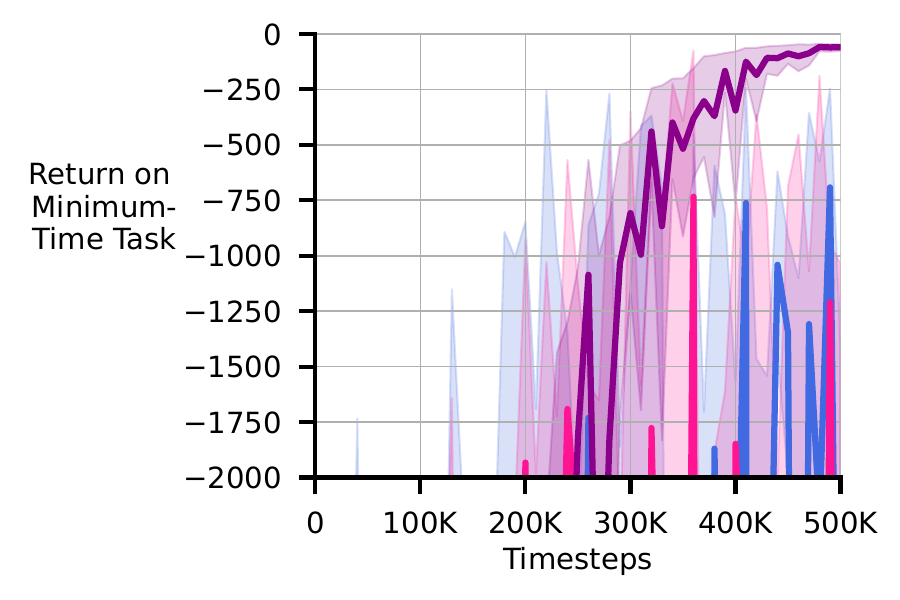}
        \caption{Minimum-Time task}
        \label{fig:vt_hard_eval_across_tasks}
    \end{subfigure}
    \caption{Evaluating the performance of all formulations across the different specifications of Reacher-Easy (Fig. \ref{fig:ar_easy_eval_across_tasks}, \ref{fig:ftl_easy_eval_across_tasks}, \ref{fig:vt_easy_eval_across_tasks}) and Reacher-Hard (Fig. \ref{fig:ar_hard_eval_across_tasks}, \ref{fig:ftl_hard_eval_across_tasks}, \ref{fig:vt_hard_eval_across_tasks}). Each solid learning is an average of 30 independent runs. The shaded regions represent a $95\%$ confidence interval.}
    \label{fig:hard_across_tasks}
\end{figure}

\paragraph{Do the learned policies achieve superior performance in their own formulations?}
We evaluate policies trained using contact reward and minimum-time formulations on the guiding reward task by assessing policies trained on other formulations every 10K timesteps. 
The minimum-time formulation performs comparably to the guiding reward formulation on Reacher-Easy (Fig. \ref{fig:ar_easy_eval_across_tasks}) but surpasses it on Reacher-Hard (Fig. \ref{fig:ar_hard_eval_across_tasks}). 
While a distinct hierarchy exists in terms of optimality  of minimum-time policies, the same is not true for policies trained with guiding rewards.

We perform a similar evaluation on other formulations.
While the contact reward policy performs on par or slightly better than the guiding reward policy, it still lags behind the minimum-time policy (Fig. \ref{fig:ftl_easy_eval_across_tasks} and \ref{fig:ftl_hard_eval_across_tasks}).
In the minimum-time task, both the guiding reward and contact reward formulations perform poorly (Fig. \ref{fig:vt_easy_eval_across_tasks} and \ref{fig:vt_hard_eval_across_tasks}). 
Notably, the agents trained by the guiding reward and contact reward formulations struggle to achieve near-zero velocity at the goal state, which is necessary for completing episodes in the minimum-time formulation. 

\section{Does Choice of Time Limit Impact Success in Minimum-Time Tasks?}
\label{sec:hit_rate}
While specifying tasks in the minimum-time formulation is straightforward, achieving successful learning can be challenging with complex real-world problems.
One factor that impacts the difficulty of a minimum-time task is the exploration strategy.
An implicit way of controlling exploration in RL is to invoke a environment reset after a certain time limit.
Many benchmark tasks in RL, such as OpenAI Gym \citep{brockman2016openai} and DeepMind Control Suite \citep{tassa2018deepmind}, including existing minimum-time tasks, such as Mountain Car \citep{sutton2018reinforcement}, treat time limits as an intrinsic property of their task formulations. 
They often use fixed time limits to separate episodes and improve exploration efficiency. 
Intuitively, a large time limit increases the chances of reaching the target since the agent has more time to explore the environment. 
However, a large time limit can hurt exploration if the agent gets stuck in uninformative states due to a sub-optimal policy. 

\begin{figure*}[htp]
\centering
    \begin{subfigure}{.25\textwidth}
        \includegraphics[width=\columnwidth]{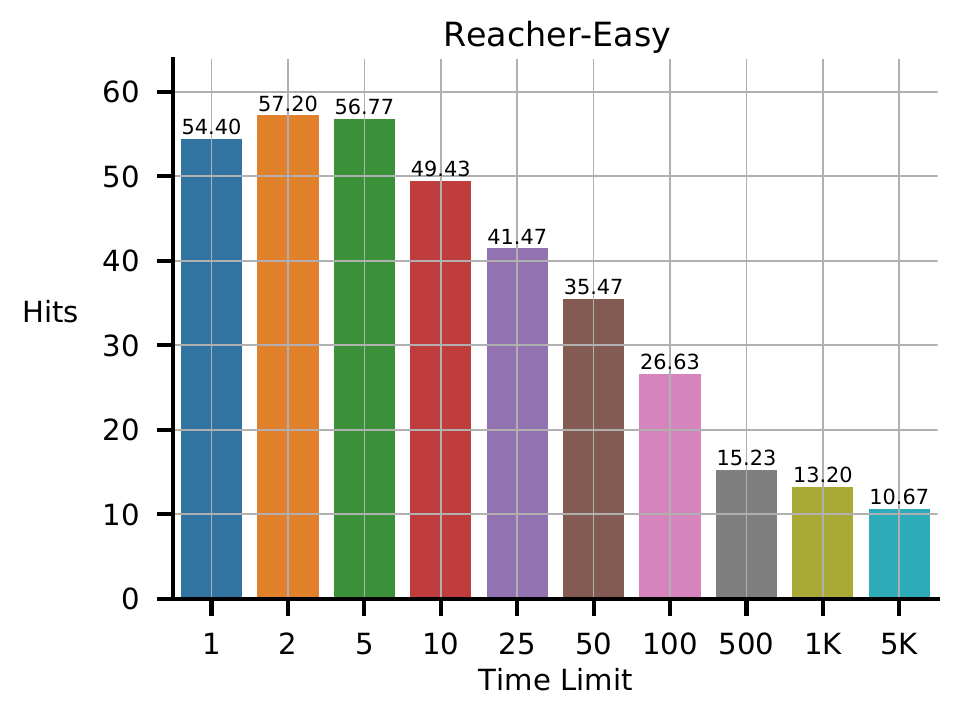}
        \label{fig:reacher_easy_init_policy_test}
    \end{subfigure}
        \begin{subfigure}{.25\textwidth}
        \includegraphics[width=\columnwidth]{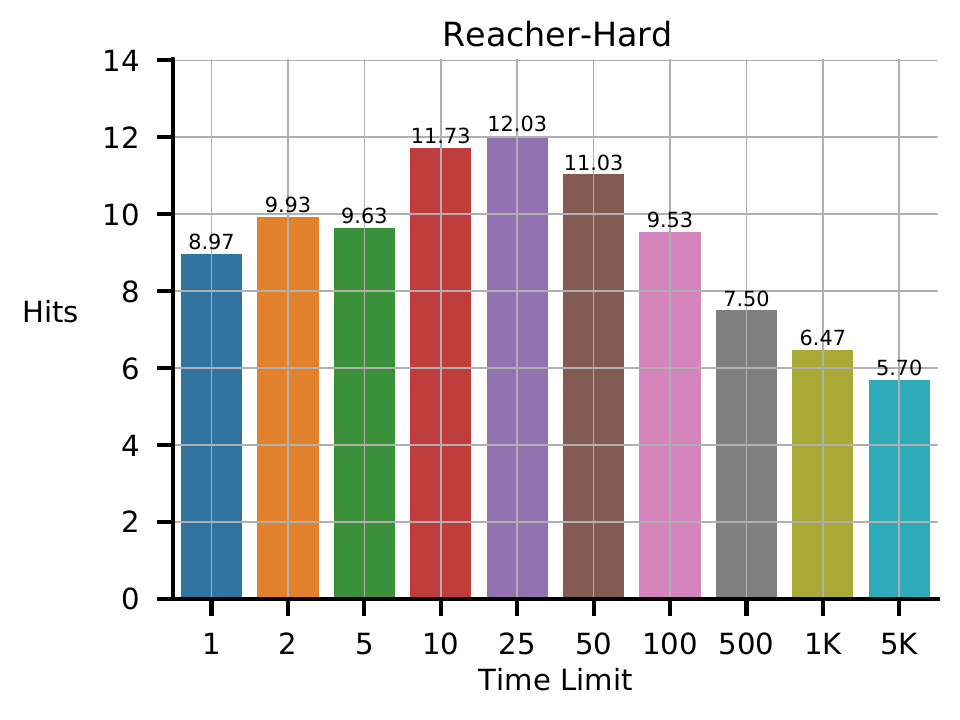}
        \label{fig:reacher_hard_init_policy_test}
    \end{subfigure}
    \begin{subfigure}{.25\textwidth}
        \includegraphics[width=\columnwidth]{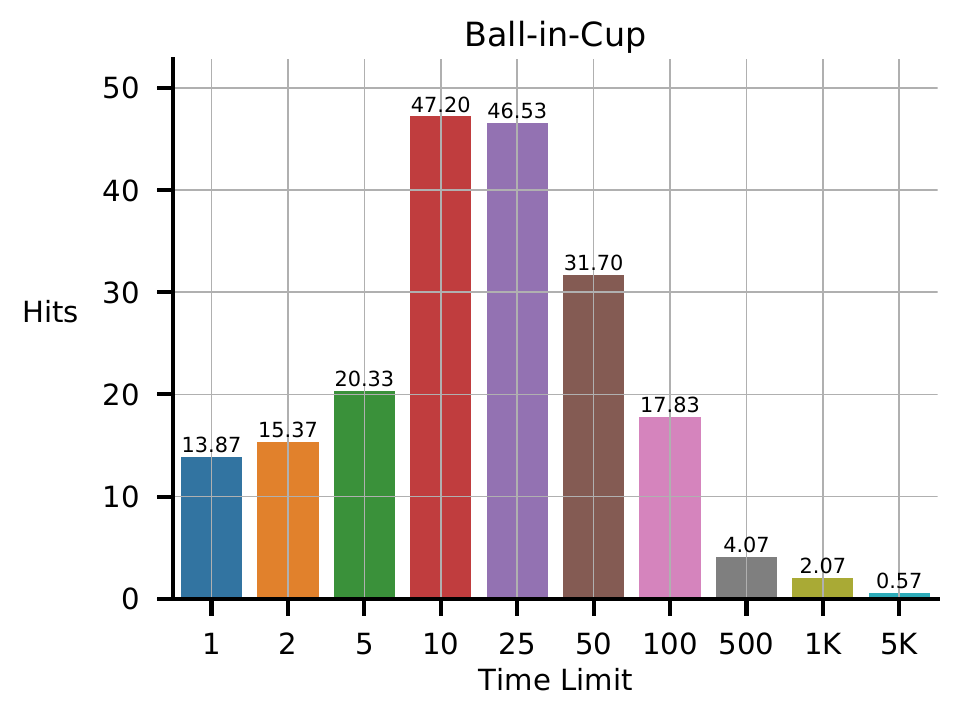}
        \label{fig:ball_in_cup_init_policy_test}
    \end{subfigure}
    \vspace{-20pt}
    \caption{A histogram plot of the choice of time limits versus the number of target hits, that is, the number of times the agent reaches the goal state using an initial policy within 20K timesteps.}
    \label{fig:simulation_init_policy_test}
\end{figure*}

To investigate the effect of the time limit on the frequency of reaching the goal state, we recorded the number of \emph{target hits}, that is, times the agent reaches the goal state using an initial gaussian policy $\mathcal{N}(0,1)$ within $20K$ timesteps. We plot the number of target hits versus the choice of time limits for a minimum-time variant of \emph{Ball-in-Cup} \citep{tassa2018deepmind} and Reacher in Fig. \ref{fig:simulation_init_policy_test}, revealing that time limits indeed affect an agent's exploration substantially.

\begin{figure*}[htp]
\centering
    \begin{subfigure}{.24\textwidth}
        \includegraphics[width=\columnwidth]{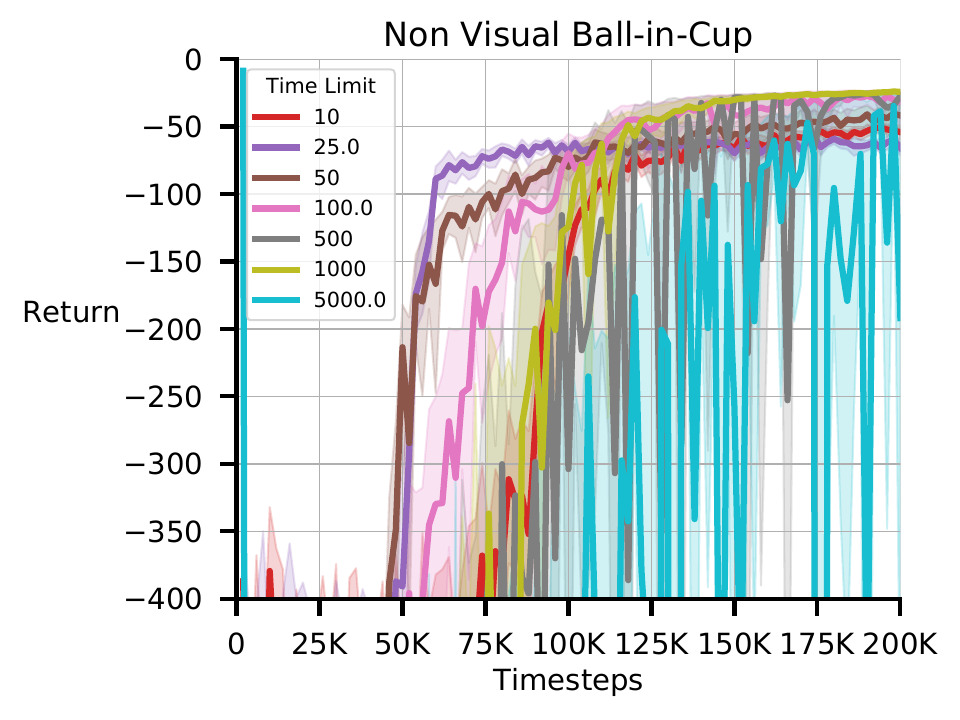}
        \label{fig:non_ball_in_cup_lc}
    \end{subfigure}
    \begin{subfigure}{.24\textwidth}
        \includegraphics[width=\columnwidth]{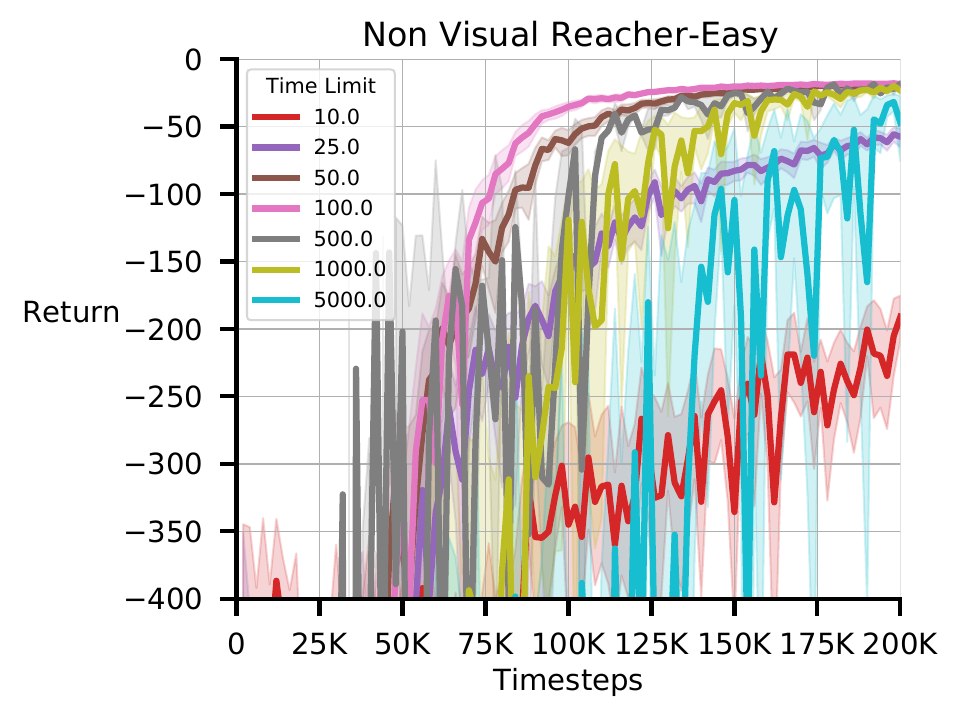}
        \label{fig:non_reacher_easy_lc}
    \end{subfigure}
    \begin{subfigure}{.24\textwidth}
        \includegraphics[width=\columnwidth]{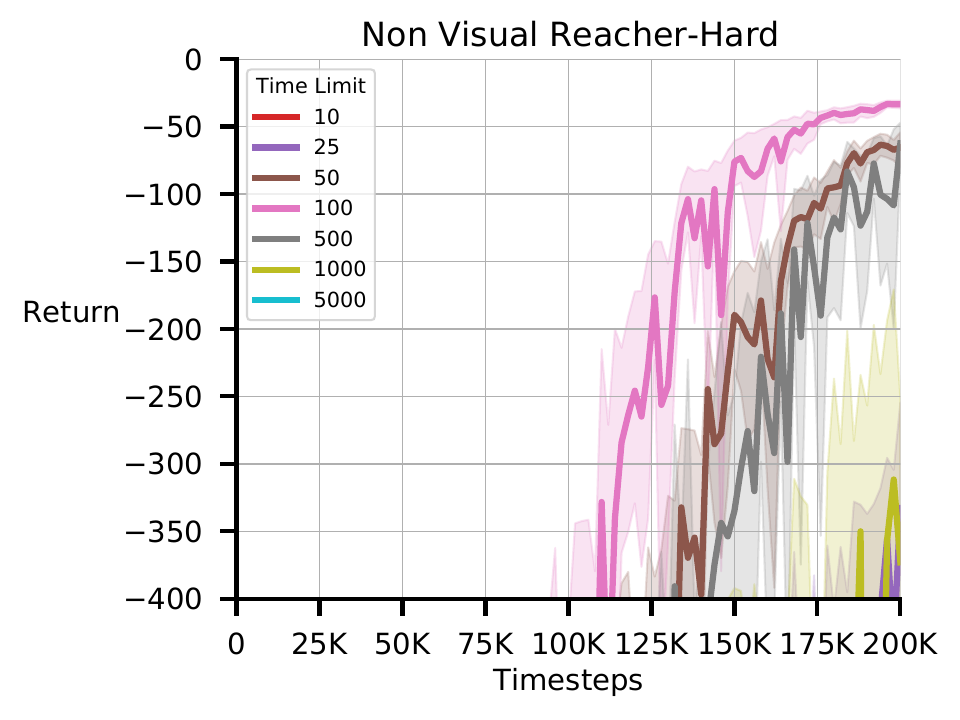}
        \label{fig:non_reacher_hard_lc}
    \end{subfigure}
    \begin{subfigure}{.24\textwidth}
        \includegraphics[width=\columnwidth]{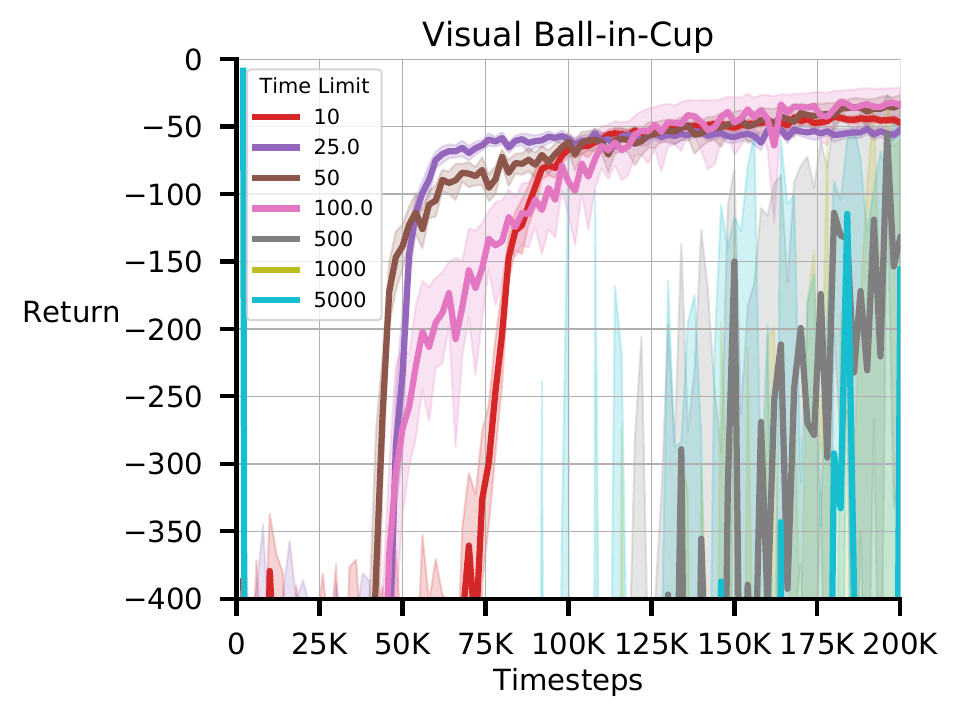}
        \label{fig:visual_ball_in_cup_lc}
    \end{subfigure}
    \vspace{-20pt}
    \caption{Learning performance of SAC on three non-visual and one visual task in simulation for multiple choices of timeout. Each solid curve is averaged over 30 independent runs. The shaded regions represent a 95\% confidence interval. }
    \label{fig:sim_exp_lc}
\end{figure*}

We used the Non-Visual and Visual Ball-in-Cup, Non-Visual Reacher-Easy, and Non-Visual Reacher-Hard tasks to test whether the number of target hits affects training performance. 
The learning curves of each task across multiple time limits shown in Fig. \ref{fig:sim_exp_lc} establish a clear relationship between the learning performance and the number of target hits.
See Appendix \ref{sec:appendix_cnn} for training details.

\paragraph{Timeout is a solution parameter} While timeouts can indeed help exploration in RL, depending on the specific context and implementation, it is oft-ignored as a parameter in task specification.
Our results \emph{confirm} that the choice of the time limit can substantially affect overall learning performance, as there is a direct correlation between the frequency of reaching the goal state and the final learning performance. 
We suggest that \emph{time limits should be treated as a tunable solution parameter that a learning system can tweak} rather than a fixed parameter inherent to the problem. 
Since our definition of an episode is independent of the choice of the time limit in the minimum-time formulation, we can safely modify the time limit of a task without altering the problem. 

\section{Learning Vision-Based Minimum-Time Tasks on Real Robots} 
\label{sec:test_guidelines}
As demonstrated in simulation, we aim to show that the initial policy's performance can help predict if a minimum-time task can be learned swiftly and consistently with real robots as well.
We employ a carefully tuned setup to showcase learning using the minimum-time formulation on four held-out robot tasks.
After analyzing our simulation results, we consider that achieving an average of $10$ target hits per $20K$ steps is sufficient for successful learning.
With the hyper-parameters we have selected, including a mini-batch size of $256$ and a replay buffer capacity of $100K$, we estimate that approximately one out of every $8$ mini-batch updates would involve a transition sample featuring the goal state. 
The replay buffer would contain at least 50+ diverse target hits. 

Using this heuristic, we aim to choose a time limit for the robot task such that the number of target hits exceeds $10$ within $20K$ timesteps using the initial policy. 
Typically, an action cycle time of $10-125Hz$ is commonly utilized with robot control. 
In this scenario, collecting $20K$ samples would take utmost 35 minutes, which is feasible for real-world experimentation.
Once an appropriate time limit is chosen, we proceed with training the task using an asynchronous implementation of the SAC algorithm and the hyper-parameters specified in Appendix \ref{sec:appendix_A}
Note that as per the minimum-time formulation, a reward of $-1$ is assigned at every step for all robot tasks until the episode ends.

\textbf{Create-Reacher} We modified the guiding reward task introduced by \cite{wang2023real} into a minimum-time task, as illustrated in Fig. \ref{fig:create_overlay}. 
To facilitate image-based tasks and onboard inference, we equipped a Depstech 4K camera and a Jetson Nano 4GB computer. 
The objective of this task is to swiftly reach one of the green sticky notes.
Each episode terminates when the green sticky note occupies at least $20\%$ of the camera image.

\textbf{UR5-VisualReacher} This task is a minimum-time adaptation of the guiding reward task proposed by \cite{yuan2022asynchronous} .
The objective is to move the robot's fingertip to the target (red blob) on the screen as fast as possible (Fig. \ref{fig:ur5_overlay}). 
The episode terminates once the target occupies more than $1.5\%$ pixels of the current image. 
The target remains unchanged until the arm reaches it. 
Upon termination, we reset the arm to a predefined posture and generate new random targets.

\textbf{Vector-ChargerDetector} We propose a novel vision-based goal-reaching task involving a low-cost mobile robot called Anki Vector (Fig. \ref{fig:vector_overlay}). 
The agent sends actions in the form of velocity commands to the wheels of the robot every $100ms$ over WiFi. 
The episode terminates when the charger symbol is centred and occupies roughly $25\%$ of the image. 
During a reset, Vector first moves backwards and reorients itself in a random direction. 

\textbf{Franka-VisualReacher}
We also introduce another novel task involving a 7-DOF robot arm, Franka Emika Panda. 
The task is to move the wrist-mounted camera close to a randomly placed bean bag on the table (see Fig. \ref{fig:franka_overlay}). 
Every episode, we place the arm and the bean bag in random positions.
The agent controls the arm by sending velocity commands for each joint at $25Hz$.
The episode ends when the bean bag covers more than 12\% of the image captured by the camera. 
We use an adaptation of the code from \cite{karimi2023dynamic} to set up this task.

\begin{table}[htp]
  \begin{center}
    \begin{adjustbox}{width=0.8\textwidth}
        \begin{tabular}{|c|c|c|c|c|c|c|} 
            \hline
              \textbf{Task} & \begin{tabular}{@{}c@{}} \textbf{Time} \\ \textbf{Limit} \end{tabular}&  \begin{tabular}{@{}c@{}} \textbf{Average} \\ \textbf{Target} \\ \textbf{Hits} \end{tabular} & \begin{tabular}{@{}c@{}} \textbf{Training} \\ \textbf{(Steps)} \end{tabular} & \begin{tabular}{@{}c@{}} \textbf{Control} \\ \textbf{Frequency} \\ \textbf{(Hz)} \end{tabular} & \begin{tabular}{@{}c@{}}\textbf{Initial Hits} \\ \textbf{Estimation} \\ \textbf{Time} \textbf{(mins)} \end{tabular} & \begin{tabular}{@{}c@{}}\textbf{Robot} \\ \textbf{Experience} \\ \textbf{Time} \textbf{(hours)} \end{tabular}\\ \hline
            \begin{tabular}{@{}c@{}} \textbf{Franka} \\ \textbf{VisualReacher} \\ \end{tabular} & \begin{tabular}{@{}c@{}c@{}}75 (3s) \\ 150 (6s) \\ 750 (30s) \end{tabular} & \begin{tabular}{@{}c@{}c@{}}$12.6 \pm 1.0$ \\ $13.0 \pm 2.08$ \\ $
    7.8 \pm 0.87$ \end{tabular} & 60k & 25 & 13.3 & $1.1\bar{1}$ \\ \hline 
            \begin{tabular}{@{}c@{}} \textbf{Create} \\ \textbf{Reacher} \end{tabular} & 333 (15s) & $18.8 \pm 3.92$  & 100k & $22.2\bar{2}$ & $15$ & $1.25$   \\ \hline 
            \begin{tabular}{@{}c@{}} \textbf{UR5} \\ \textbf{VisualReacher} \\ \end{tabular}  & 150 (6s) & $14.4 \pm 2.5$  & 100k & 25 & 13.3 & $1.1\bar{1}$ \\ \hline 
            \begin{tabular}{@{}c@{}} \textbf{Vector} \\ \textbf{ChargerDetector} \\ \end{tabular}  & 300 (30s) & $11 \pm 1.1$  & 160k & 10 & $33.3\bar{3}$ & $4.4\bar{4}$ \\ \hline 
        \end{tabular}
    \end{adjustbox}
  \end{center}
  \caption{Robot experiment setup. The average target hits are calculated over five independent runs using the initial Gaussian policy $\mathcal{N}(0, 1)$ spanning $20K$ timesteps for each task. Note that robot experience time does not include time required to reset the environment, charging the batteries, etc.}
  \vspace{-10pt}
  \label{tab:robot_timeouts}
\end{table}

\begin{figure}[htp]
\centering
    \begin{subfigure}{.235\columnwidth}
        \includegraphics[width=\columnwidth]{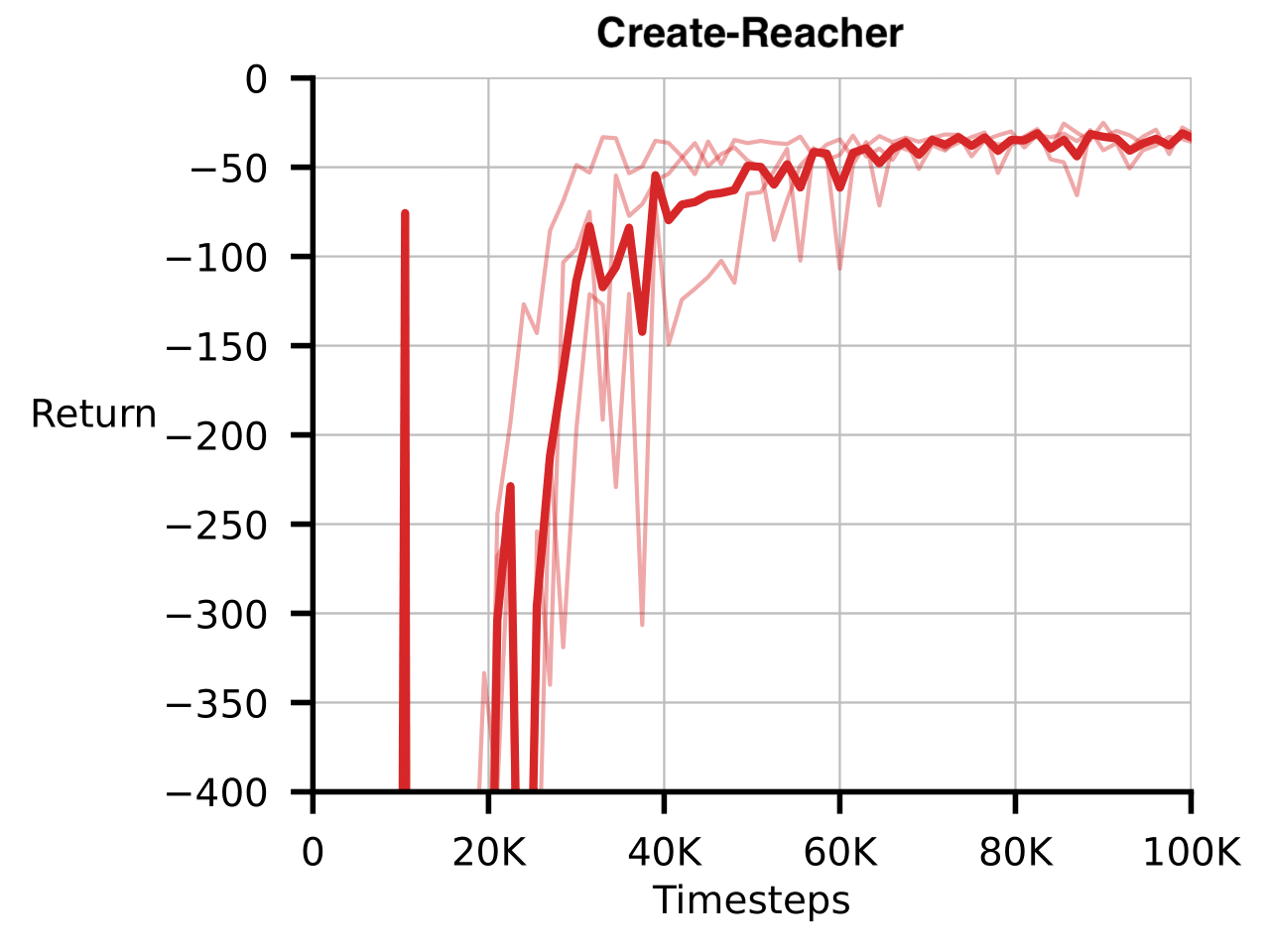}
        \label{fig:create2_visual_reacher_lc}
    \end{subfigure}
    \begin{subfigure}{.235\columnwidth}
        \includegraphics[width=\columnwidth]{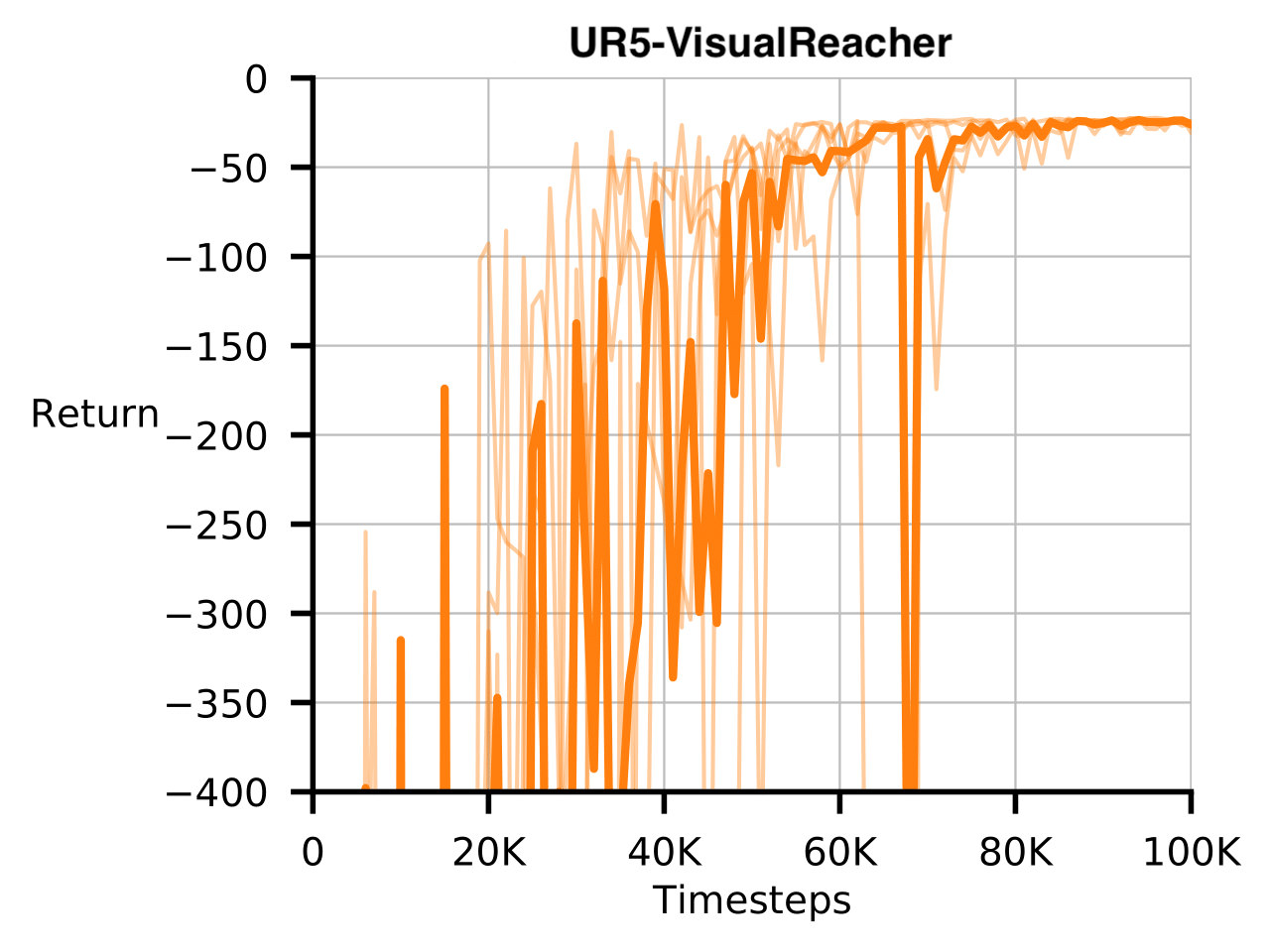}
        \label{fig:ur5_lc}
    \end{subfigure}
    \begin{subfigure}{.235\columnwidth}
        \includegraphics[width=\columnwidth]{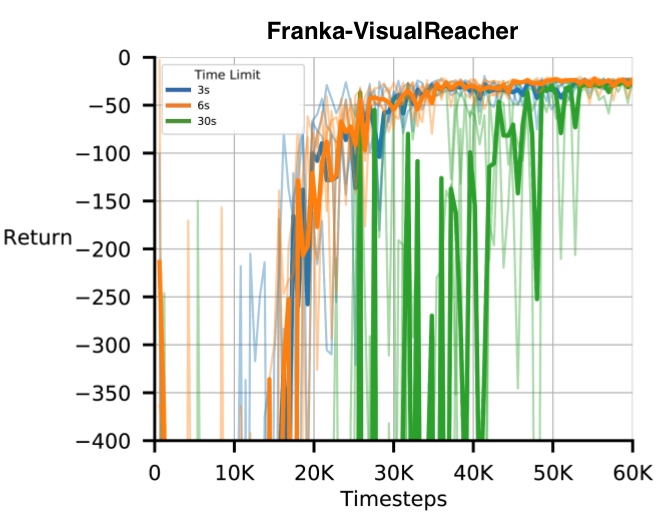}
        \label{fig:franka_lc}
    \end{subfigure}
    \begin{subfigure}{.235\columnwidth}
        \includegraphics[width=\columnwidth]{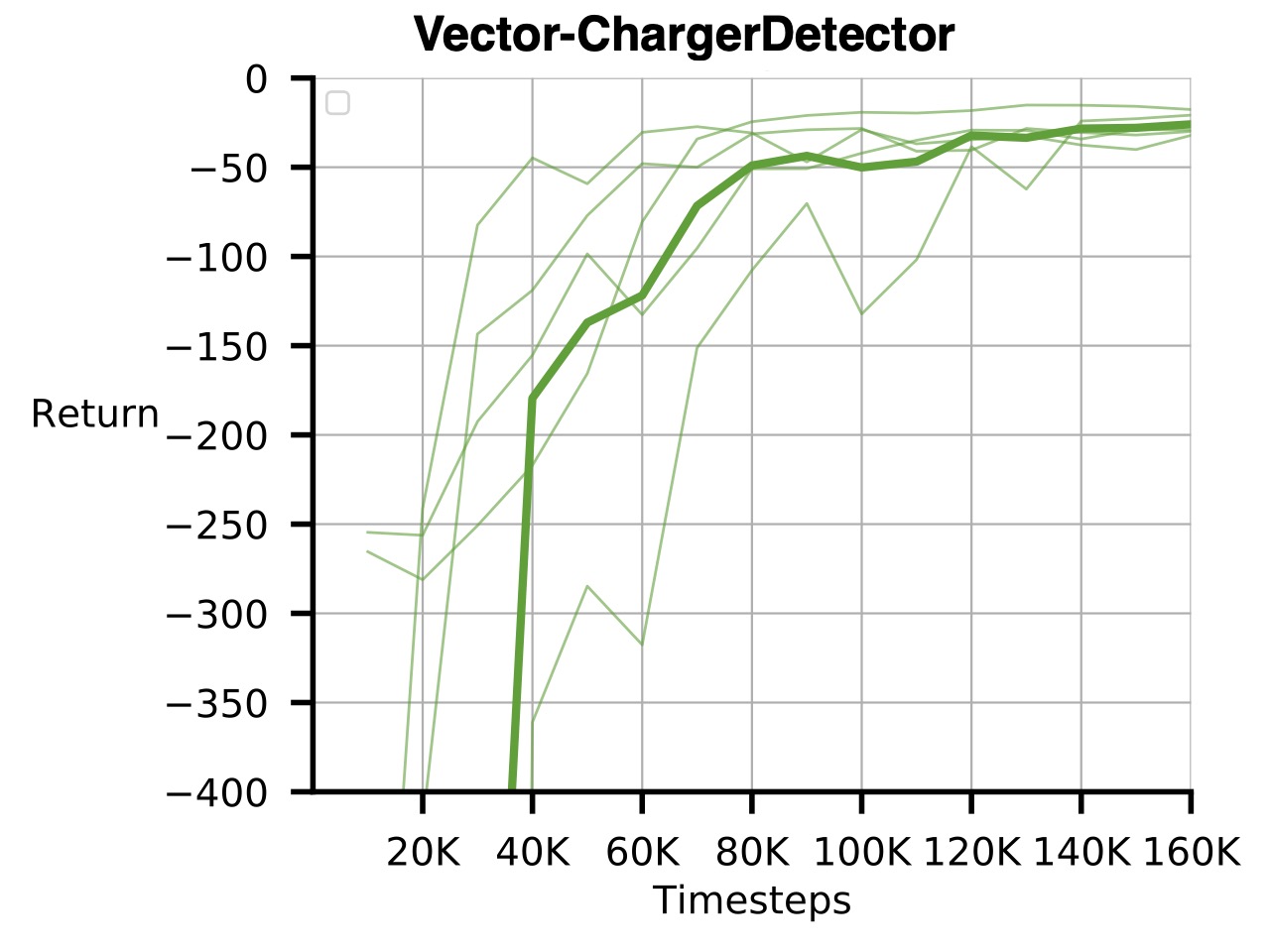}
        \label{fig:vector_lc}
    \end{subfigure}
    \vspace{-20pt}
    \caption{Learning performance of SAC on four vision-based policy learning tasks with robots. Each thin learning curve here represents an independent run. The bold learning curves of the same color are the average of all independent runs.}
    \label{fig:real_robot_lc}
\end{figure}

While seemingly simple, these tasks pose significant complexity for model-free learning algorithms aiming to learn vision-based policies from scratch within a reasonable timeframe. 
The initial random policy performs poorly, warranting its exclusion from Fig. \ref{fig:real_robot_lc}. Real robot task data collection is slow and costly, leading us to test only three time limits for Franka-VisualReacher and select one heuristically for other tasks (Table \ref{tab:robot_timeouts}). 
Five runs were conducted for each task. 
Utilizing the \emph{ReLoD} system \citep{wang2023real} is essential for effective real-time learning. Franka-VisualReacher and Vector-ChargerDetector achieve sufficient target hits, while Create-Reacher and UR5-VisualReacher surpass this threshold. 
We anticipate SAC agents can effectively learn policies for all tasks, which is supported by the successful learning curves depicted in Fig. \ref{fig:real_robot_lc}. 
This validates our observation that the initial policy's performance is indicative of successful learning on minimum-time tasks.

\paragraph{Using a curriculum to train a robust policy}
Using a hand-crafted curriculum in RL involves designing a structured sequence of tasks to accelerate an agent's learning. 
Beginning with simpler tasks and gradually introducing more challenging ones accelerates skill development and knowledge transfer. 
It's possible to seamlessly integrate a curriculum with minimum-time tasks, where the acceptable threshold for reaching a goal state can be progressively reduced over time.
We present a single demonstration featuring the UR5-VisualReacher and Franka-VisualReacher tasks, which were trained with a curriculum over $200K$ timesteps, and we showcase their performance in our \href{https://drive.google.com/file/d/1O8D3oCWq5xf2hi1JOlMBbs6W1ClrvUFb/view?usp=sharing}{video}\footnote{\url{https://drive.google.com/file/d/1O8D3oCWq5xf2hi1JOlMBbs6W1ClrvUFb/view?usp=sharing}}

\paragraph{Evaluation of final learned behaviors on the robot arms}
\begin{figure}[htp]
    \centering
    \vspace{-5pt}
    \begin{subfigure}{.235\columnwidth}
        \includegraphics[width=\columnwidth]{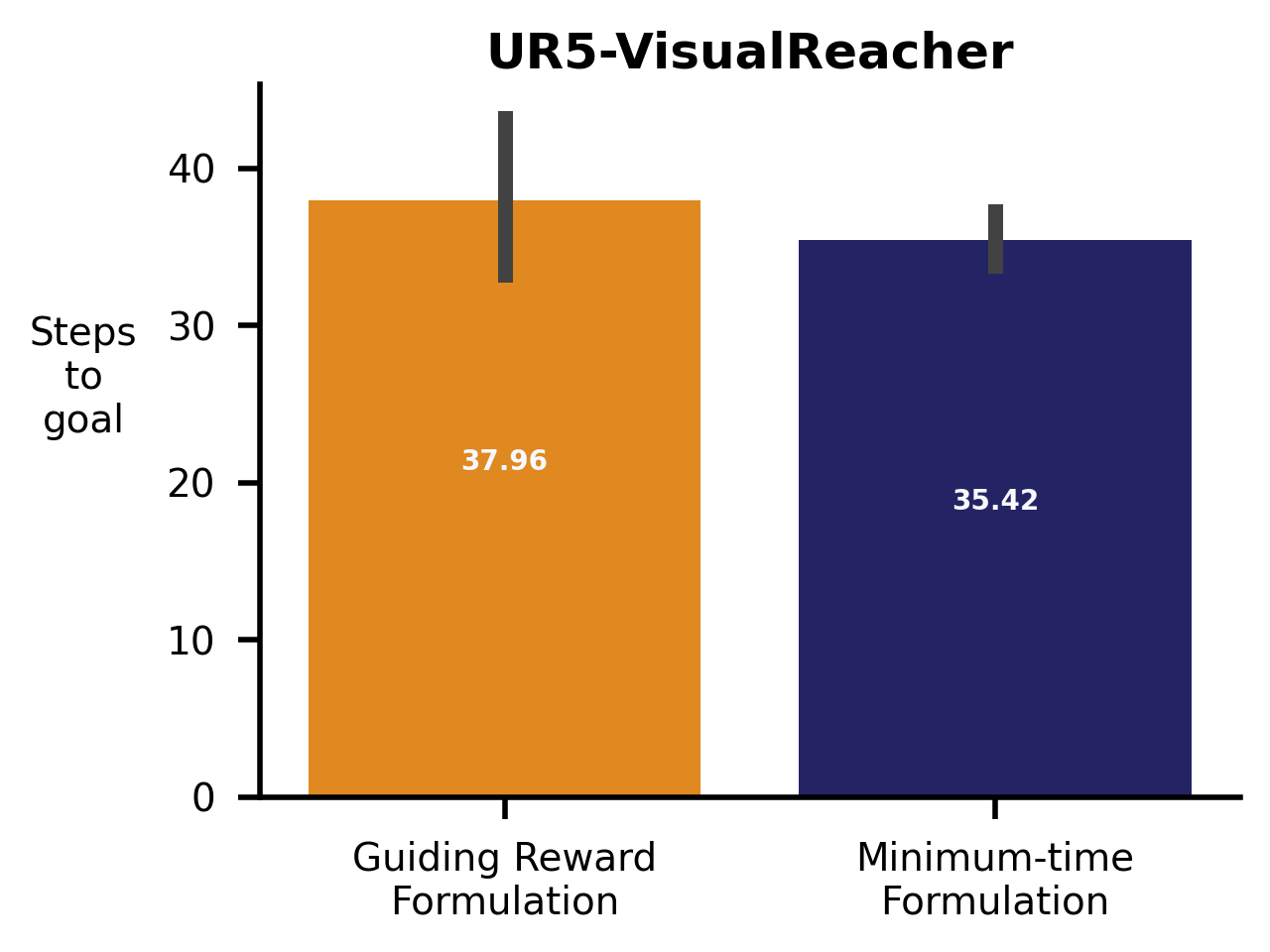}
        \caption{Steps to goal plot for UR5-VisualReacher}
        \label{fig:ur5_steps_to_goal}
    \end{subfigure}
    \begin{subfigure}{.235\columnwidth}
        \includegraphics[width=\columnwidth]{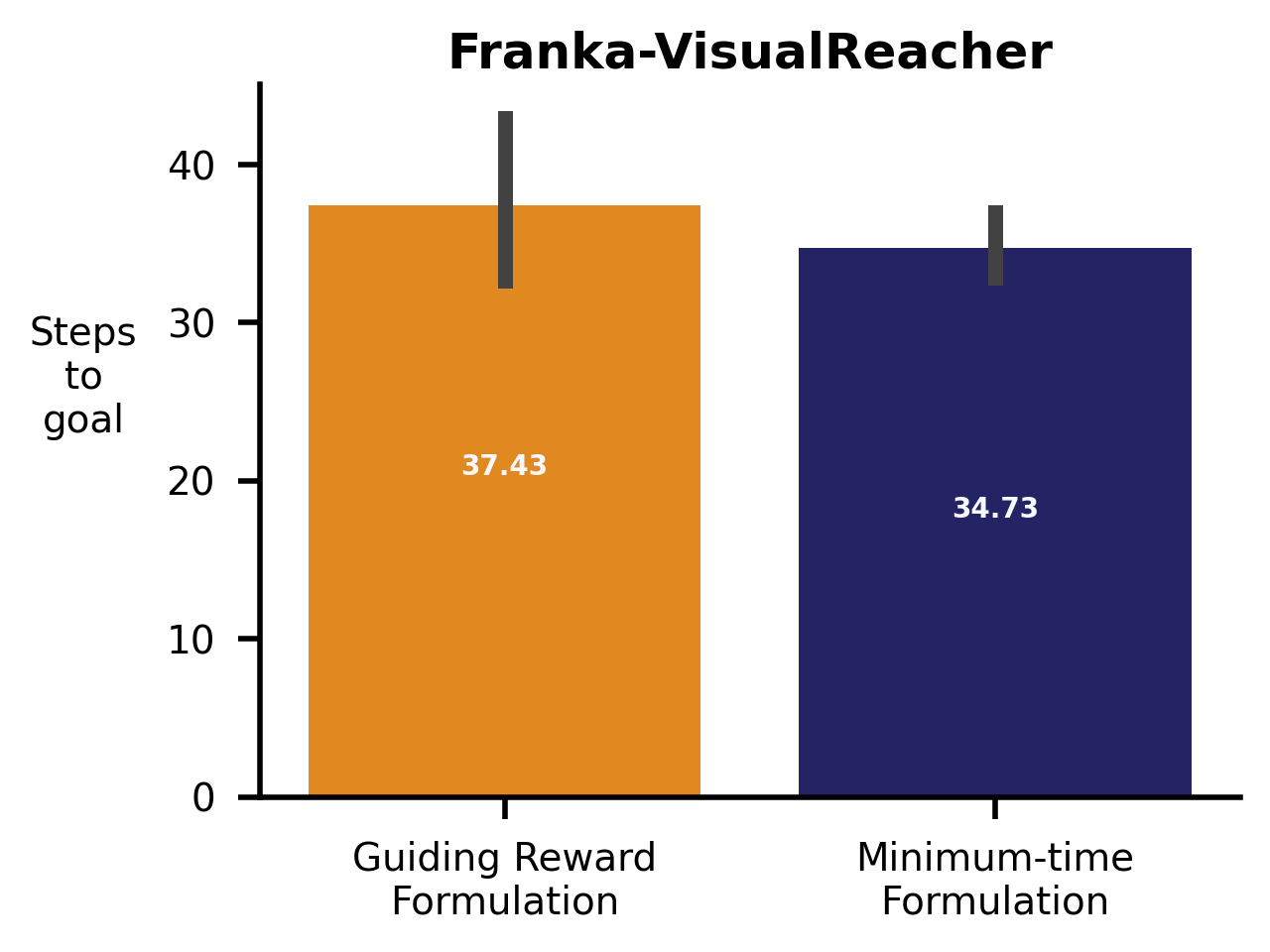}
        \caption{Steps to goal plot for Franka-VisualReacher}
        \label{fig:franka_steps_to_goal}
    \end{subfigure}
        \begin{subfigure}{.235\columnwidth}
        \includegraphics[width=\columnwidth]{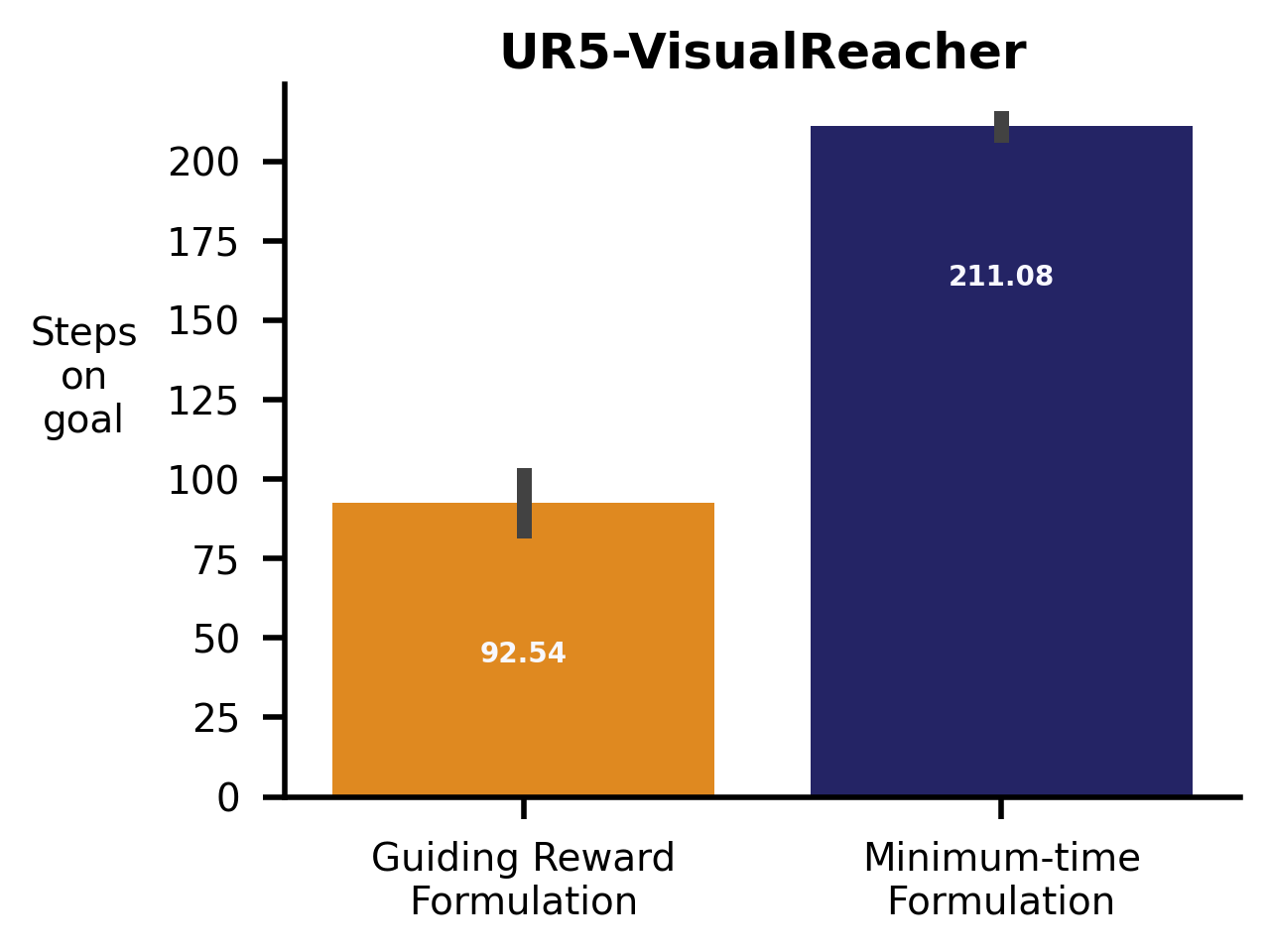}
        \caption{Steps on goal plot for UR5-VisualReacher}
        \label{fig:ur5_steps_on_goal}
    \end{subfigure}
    \begin{subfigure}{.235\columnwidth}
        \includegraphics[width=\columnwidth]{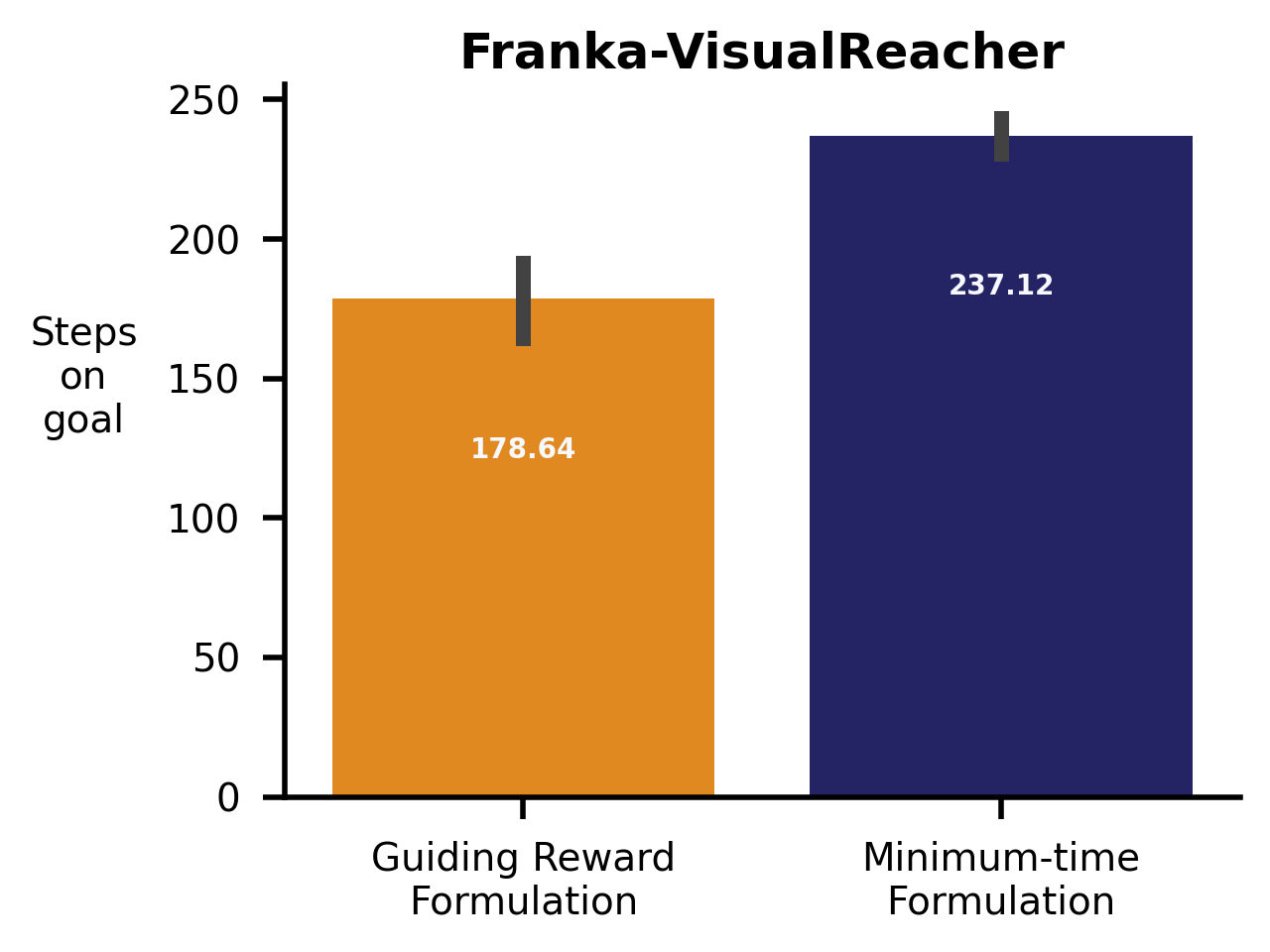}
        \caption{Steps on goal plot for Franka-VisualReacher}
        \label{fig:franka_steps_on_goal}
    \end{subfigure}
    \caption{Comparison of the guiding reward and minimum-time formulations of the reaching task using the number of steps it requires to reach the goal (\ref{fig:ur5_steps_to_goal}, \ref{fig:franka_steps_to_goal}) and the number of steps it stays on target upon reaching it (\ref{fig:ur5_steps_on_goal}, \ref{fig:franka_steps_on_goal}).  In bar charts \ref{fig:ur5_steps_to_goal} \& \ref{fig:franka_steps_to_goal}, a lower value signifies better performance. In \ref{fig:ur5_steps_on_goal} \& \ref{fig:franka_steps_on_goal}, a higher value signifies better performance.}
    \label{fig:robot_stg}
\end{figure}

We evaluate UR5-VisualReacher and Franka-VisualReacher policies under minimum-time and guiding reward formulations, akin to Section \ref{sec:eval_final}, using steps to goal and steps on goal metrics. The guiding reward task has been previously used in \cite{yuan2022asynchronous} and \cite{grooten24madi} and is detailed in Appendix \ref{sec:appendix_robot}.
We conduct $2500$ episodes lasting $500$ timesteps each ($1.25M$ steps total) for UR5-VisualReacher and $1000$ episodes lasting $1000$ timesteps each ($1M$ steps total) for Franka-VisualReacher. Analysis of Figure \ref{fig:robot_stg} shows that minimum-time policies are marginally faster and maintain target contact longer compared to guiding reward policies, consistent with simulation results (Fig. \ref{fig:steps_to_goal}), highlighting the effectiveness of the minimum-time formulation.

\section{Conclusions}
Our research advocates for re-evaluating the use of constant negative rewards in real robot learning as a cost-effective approach to reduce the need for extensive reward engineering and achieve superior final performance. 
We showed that the minimum-time approach not only simplifies task specification but can also surprisingly lead to better policies, even when assessed using metrics designed for the guiding reward approach, as demonstrated through empirical evidence in simulation and real robots.
Through an empirical investigation of the performance of SAC on multiple complex vision-based simulated and real robotic control tasks in the minimum-time formulation, we identified that an agent could achieve successful learning performance if it can reach the goal often enough using its initial policy. 
Contrary to popular belief, we showed that it is possible to have robust, and reliable learning from scratch on complex vision-based robotic control tasks within a reasonable timeframe using only constant negative rewards. 
We established that the time limit should be a tunable solution parameter instead of a problem parameter in the minimum-time formulation.
In total, we conducted multiple independent runs in our experiments, which took nearly 200 hours of usage on real robots.
Our model-free reinforcement learning system achieved real-time learning of pixel-based control of several different kinds of physical robots from scratch.

\subsubsection*{Acknowledgments}
\label{sec:ack}
We would like to thank Richard Sutton, Matthew Taylor, Patrick Pilarski, Shivam Garg, Varshini Prakash and Bram Grooten for several insightful discussions that helped improve the quality and clarity of the paper.
We would also like to thank Homayoon Farrahi for providing the training script that helped us learn the UR5-Reacher task using a curriculum.
We are also appreciative of the computing resources provided by the Digital Research Alliance of Canada and the financial support from the CCAI Chairs program, the RLAI laboratory, Amii, and NSERC of Canada. We are also thankful to Ocado Technology and Huawei Noah’s Ark Lab for their generous donation of the UR5 and Franka Emika Panda, respectively.


\bibliography{main}

\begin{thebibliography}{}

\bibitem[Amodei et~al., 2016]{amodei2016concrete}
Amodei, D., Olah, C., Steinhardt, J., Christiano, P., Schulman, J., and Man{\'e}, D. (2016).
\newblock Concrete problems in ai safety.
\newblock {\em arXiv preprint arXiv:1606.06565}.

\bibitem[Andrychowicz et~al., 2017]{andrychowicz2017hindsight}
Andrychowicz, M., Wolski, F., Ray, A., Schneider, J., Fong, R., Welinder, P., McGrew, B., Tobin, J., Pieter~Abbeel, O., and Zaremba, W. (2017).
\newblock Hindsight experience replay.
\newblock {\em Advances in neural information processing systems}, 30.

\bibitem[Booth et~al., 2023]{booth2023perils}
Booth, S., Knox, W.~B., Shah, J., Niekum, S., Stone, P., and Allievi, A. (2023).
\newblock The perils of trial-and-error reward design: misdesign through overfitting and invalid task specifications.
\newblock In {\em Proceedings of the AAAI Conference on Artificial Intelligence}, volume~37, pages 5920--5929.

\bibitem[Brockman et~al., 2016]{brockman2016openai}
Brockman, G., Cheung, V., Pettersson, L., Schneider, J., Schulman, J., Tang, J., and Zaremba, W. (2016).
\newblock Openai gym.
\newblock {\em arXiv preprint arXiv:1606.01540}.

\bibitem[Che et~al., 2023]{che2023correcting}
Che, F., Vasan, G., and Mahmood, A.~R. (2023).
\newblock Correcting discount-factor mismatch in on-policy policy gradient methods.
\newblock In {\em International Conference on Machine Learning}, pages 4218--4240. PMLR.

\bibitem[Chui and Chen, 2012]{chui2012linear}
Chui, C.~K. and Chen, G. (2012).
\newblock {\em Linear systems and optimal control}, volume~18.
\newblock Springer Science \& Business Media.

\bibitem[Elsayed et~al., 2024]{elsayed2024revisiting}
Elsayed, M., Farrahi, H., Dangel, F., and Mahmood, A.~R. (2024).
\newblock Revisiting scalable hessian diagonal approximations for applications in reinforcement learning.
\newblock In {\em International conference on machine learning}.

\bibitem[Farrahi and Mahmood, 2023]{farrahi2023reducing}
Farrahi, H. and Mahmood, A.~R. (2023).
\newblock Reducing the cost of cycle-time tuning for real-world policy optimization.
\newblock In {\em 2023 International Joint Conference on Neural Networks (IJCNN)}, pages 1--8. IEEE.

\bibitem[Garg et~al., 2022]{garg2021alternate}
Garg, S., Tosatto, S., Pan, Y., White, M., and Mahmood, A.~R. (2022).
\newblock An alternate policy gradient estimator for softmax policies.
\newblock In {\em International Conference on Artificial Intelligence and Statistics}.

\bibitem[Grooten et~al., 2024]{grooten24madi}
Grooten, B., Tomilin, T., Vasan, G., Taylor, M.~E., Mahmood, R.~A., Fang, M., Pechenizkiy, M., and Mocanu, D.~C. (2024).
\newblock Madi: Learning to mask distractions for generalization in visual deep reinforcement learning.
\newblock In {\em AAMAS'24: 2024 International Conference on Autonomous Agents and Multiagent Systems}. International Foundation for Autonomous Agents and Multiagent Systems~….

\bibitem[Haarnoja et~al., 2018]{haarnoja2018soft}
Haarnoja, T., Zhou, A., Abbeel, P., and Levine, S. (2018).
\newblock Soft actor-critic: Off-policy maximum entropy deep reinforcement learning with a stochastic actor.
\newblock In {\em International conference on machine learning}, pages 1861--1870. PMLR.

\bibitem[Hertweck et~al., 2020]{hertweck2020simple}
Hertweck, T., Riedmiller, M., Bloesch, M., Springenberg, J.~T., Siegel, N., Wulfmeier, M., Hafner, R., and Heess, N. (2020).
\newblock Simple sensor intentions for exploration.
\newblock {\em arXiv preprint arXiv:2005.07541}.

\bibitem[Karimi et~al., 2023]{karimi2023dynamic}
Karimi, A., Jin, J., Luo, J., Mahmood, A.~R., Jagersand, M., and Tosatto, S. (2023).
\newblock Dynamic decision frequency with continuous options.
\newblock In {\em 2023 IEEE/RSJ International Conference on Intelligent Robots and Systems (IROS)}, pages 7545--7552. IEEE.

\bibitem[Knox et~al., 2023]{knox2023reward}
Knox, W.~B., Allievi, A., Banzhaf, H., Schmitt, F., and Stone, P. (2023).
\newblock Reward (mis) design for autonomous driving.
\newblock {\em Artificial Intelligence}, 316:103829.

\bibitem[Kober et~al., 2013]{kober2013reinforcement}
Kober, J., Bagnell, J.~A., and Peters, J. (2013).
\newblock Reinforcement learning in robotics: A survey.
\newblock {\em The International Journal of Robotics Research}, 32(11):1238--1274.

\bibitem[Korenkevych et~al., 2019]{korenkevych2019autoregressive}
Korenkevych, D., Mahmood, A.~R., Vasan, G., and Bergstra, J. (2019).
\newblock Autoregressive policies for continuous control deep reinforcement learning.
\newblock In {\em Proceedings of the 28th International Joint Conference on Artificial Intelligence}, pages 2754--2762.

\bibitem[Lan et~al., 2022]{lan2022model}
Lan, Q., Tosatto, S., Farrahi, H., and Mahmood, R. (2022).
\newblock Model-free policy learning with reward gradients.
\newblock In {\em International Conference on Artificial Intelligence and Statistics}, pages 4217--4234. PMLR.

\bibitem[Laskin et~al., 2020]{laskin2020reinforcement}
Laskin, M., Lee, K., Stooke, A., Pinto, L., Abbeel, P., and Srinivas, A. (2020).
\newblock Reinforcement learning with augmented data.
\newblock {\em Advances in neural information processing systems}, 33:19884--19895.

\bibitem[Lee et~al., 2019]{lee2019making}
Lee, M.~A., Zhu, Y., Srinivasan, K., Shah, P., Savarese, S., Fei-Fei, L., Garg, A., and Bohg, J. (2019).
\newblock Making sense of vision and touch: Self-supervised learning of multimodal representations for contact-rich tasks.
\newblock In {\em 2019 International Conference on Robotics and Automation (ICRA)}, pages 8943--8950. IEEE.

\bibitem[Levine et~al., 2016]{levine2016end}
Levine, S., Finn, C., Darrell, T., and Abbeel, P. (2016).
\newblock End-to-end training of deep visuomotor policies.
\newblock {\em The Journal of Machine Learning Research}, 17(1):1334--1373.

\bibitem[Ma et~al., 2023]{ma2023eureka}
Ma, Y.~J., Liang, W., Wang, G., Huang, D.-A., Bastani, O., Jayaraman, D., Zhu, Y., Fan, L., and Anandkumar, A. (2023).
\newblock Eureka: Human-level reward design via coding large language models.
\newblock {\em arXiv preprint arXiv:2310.12931}.

\bibitem[Mahmood et~al., 2018]{mahmood2018benchmarking}
Mahmood, A.~R., Korenkevych, D., Vasan, G., Ma, W., and Bergstra, J. (2018).
\newblock Benchmarking reinforcement learning algorithms on real-world robots.
\newblock In {\em Conference on robot learning}, pages 561--591. PMLR.

\bibitem[Mataric, 1994]{mataric1994reward}
Mataric, M.~J. (1994).
\newblock Reward functions for accelerated learning.
\newblock In {\em Machine learning proceedings 1994}, pages 181--189. Elsevier.

\bibitem[Nair et~al., 2018]{nair2018visual}
Nair, A.~V., Pong, V., Dalal, M., Bahl, S., Lin, S., and Levine, S. (2018).
\newblock Visual reinforcement learning with imagined goals.
\newblock {\em Advances in neural information processing systems}, 31.

\bibitem[Ng et~al., 1999]{ng1999policy}
Ng, A.~Y., Harada, D., and Russell, S. (1999).
\newblock Policy invariance under reward transformations: Theory and application to reward shaping.
\newblock In {\em Icml}, volume~99, pages 278--287. Citeseer.

\bibitem[Penicka et~al., 2022]{penicka2022learning}
Penicka, R., Song, Y., Kaufmann, E., and Scaramuzza, D. (2022).
\newblock Learning minimum-time flight in cluttered environments.
\newblock {\em IEEE Robotics and Automation Letters}, 7(3):7209--7216.

\bibitem[Riedmiller et~al., 2018]{riedmiller2018learning}
Riedmiller, M., Hafner, R., Lampe, T., Neunert, M., Degrave, J., Wiele, T., Mnih, V., Heess, N., and Springenberg, J.~T. (2018).
\newblock Learning by playing solving sparse reward tasks from scratch.
\newblock In {\em International conference on machine learning}, pages 4344--4353. PMLR.

\bibitem[Rocamonde et~al., 2023]{rocamonde2023vision}
Rocamonde, J., Montesinos, V., Nava, E., Perez, E., and Lindner, D. (2023).
\newblock Vision-language models are zero-shot reward models for reinforcement learning.
\newblock {\em arXiv preprint arXiv:2310.12921}.

\bibitem[Sutton and Barto, 2018]{sutton2018reinforcement}
Sutton, R.~S. and Barto, A.~G. (2018).
\newblock {\em Reinforcement learning: An introduction}.
\newblock MIT press.

\bibitem[Tassa et~al., 2018]{tassa2018deepmind}
Tassa, Y., Doron, Y., Muldal, A., Erez, T., Li, Y., Casas, D. d.~L., Budden, D., Abdolmaleki, A., Merel, J., Lefrancq, A., et~al. (2018).
\newblock Deepmind control suite.
\newblock {\em arXiv preprint arXiv:1801.00690}.

\bibitem[Vasan and Pilarski, 2017]{vasan2017learning}
Vasan, G. and Pilarski, P.~M. (2017).
\newblock Learning from demonstration: Teaching a myoelectric prosthesis with an intact limb via reinforcement learning.
\newblock In {\em 2017 International Conference on Rehabilitation Robotics (ICORR)}, pages 1457--1464. IEEE.

\bibitem[Wang et~al., 2024]{wang2024rl}
Wang, Y., Sun, Z., Zhang, J., Xian, Z., Biyik, E., Held, D., and Erickson, Z. (2024).
\newblock Rl-vlm-f: Reinforcement learning from vision language foundation model feedback.
\newblock {\em arXiv preprint arXiv:2402.03681}.

\bibitem[Wang et~al., 2023]{wang2023real}
Wang, Y., Vasan, G., and Mahmood, A.~R. (2023).
\newblock Real-time reinforcement learning for vision-based robotics utilizing local and remote computers.
\newblock In {\em 2023 IEEE International Conference on Robotics and Automation (ICRA)}, pages 9435--9441. IEEE.

\bibitem[Yuan and Mahmood, 2022]{yuan2022asynchronous}
Yuan, Y. and Mahmood, A.~R. (2022).
\newblock Asynchronous reinforcement learning for real-time control of physical robots.
\newblock In {\em 2022 International Conference on Robotics and Automation (ICRA)}, pages 5546--5552. IEEE.

\end{thebibliography}
\bibliographystyle{apalike}

\newpage
\appendix

\section{Appendix}


\subsection{Choice of Hyper-parameters}
\label{sec:appendix_A}
\begin{table}[h]
    \begin{center}
        \begin{tabular}{|c|c|c|} 
            \hline
            \textbf{Hyper-parameter} & \textbf{Values}  \\ \hline
            Replay buffer & $100K$ \\ \hline
            Actor step size & 3e-4 \\ \hline
            Critic step size & 3e-4 \\ \hline
            Entropy coefficient step size & 3e-4 \\ \hline
            Batch size & 256 \\ \hline
            Discount factor ($\gamma$) & 0.99 \\ \hline
            Update every $k$ steps & 2 \\ \hline
            Num. update epochs every $k^{th}$ step & 1 \\ \hline
            Actor MLP hidden sizes & [512 512] \\ \hline
            Critic MLP hidden sizes & [512 512] \\ \hline
            Warm-up time steps & 1000 \\ \hline
            Adam optimizer betas & [0.9, 0.999] \\ \hline
            Initial temperature & 0.1 \\ \hline
            Neural network activation & ReLU \\ \hline
        \end{tabular}
        \caption{}
        \label{tab:hp}
    \end{center}
\end{table}

In most configurations, effective learning is achieved without tuning the hyper-parameters, which indicates the reliability of the algorithm, our learning architecture, and our task setup. Note that for experiments in Sec. \ref{sec:hit_rate}, the warm-up time steps is set to 20000.

\subsection{Neural Network Architecture}
\label{sec:appendix_cnn}
\begin{wrapfigure}{r}{.5\columnwidth}
    \centering
    \vspace{-20pt}
    \centering
    \includegraphics[width=0.5\textwidth]{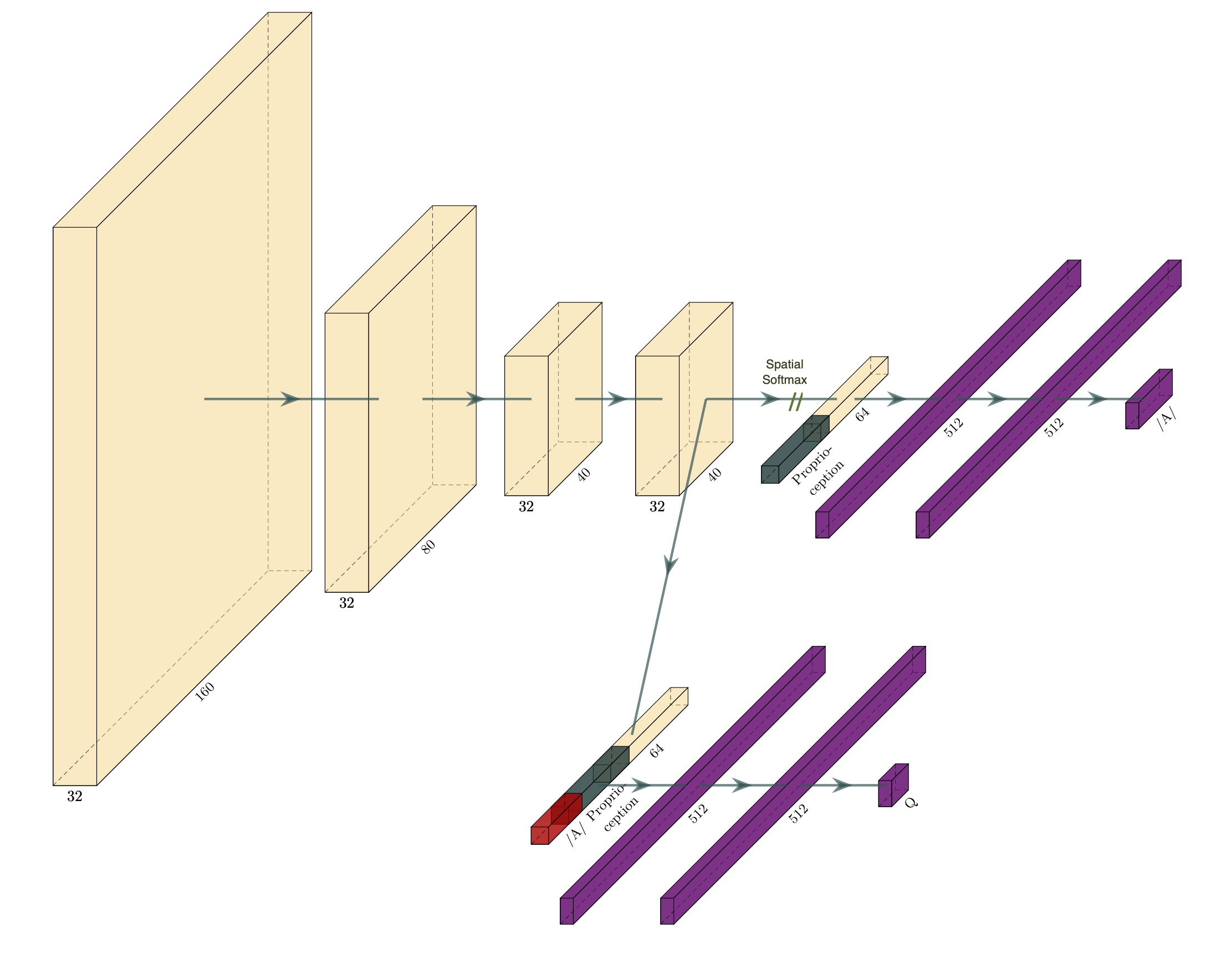}
    \caption{Our CNN architecture combines vision and proprioception information. The yellow blocks represent convolutional layers. The teal block represents proprioception vector. The magenta blocks represent fully connected layers. The red block represents the previous action.}
    \label{fig:cnn}
    \vspace{-10pt}
\end{wrapfigure}

Our convolutional neural network (CNN) architecture comprises four convolutional layers, followed by a combination of a Spatial Softmax layer and proprioception information. 
The convolutional layers have 32 output channels and 3x3 kernels, with stride of two for the first three layers and one for the last layer.
After these convolutional layers, tuse spatial-softmax \citep{levine2016end} to convert the encoding vector into soft coordinates to track the target more precisely. 
Additionally, proprioception information is concatenated with the spatial softmax features. 
The exact number of parameters depends on the input data size and task-specific requirements. 
The two MLP layers have 512 hidden units each.
All the layers except the final output layer use ReLU activation.
We use a squashed Gaussian policy, which means that samples are obtained according to:

\[
a_\theta (s, \varepsilon) = tanh(\mu_\theta(s) + \sigma_\theta(s) \odot \varepsilon), \varepsilon \sim \mathcal{N}(0, I)
\]

We also apply random cropping to augment images in mini-batches to learn more robust representations given our limited amount of observations \citep{laskin2020reinforcement}.
Note that we train the encoder only using the critic loss. 
The actor does not propagate any gradients to the shared encoder.

\subsection{Guiding Reward Specification for the UR5 and Franka Reaching Tasks}
\label{sec:appendix_robot}
The reward function is defined as follows \citep{yuan2022asynchronous}:

\begin{equation*}
    r_t= \frac{c}{hw} M_t \odot W   
\end{equation*}
Here, $c$ represents a scaling coefficient, while $h$ and $w$ denote the height and width of the image in pixels, respectively. $M_t$ is a binary mask of shape $h \times w$, identifying whether each pixel currently displays the color red. Additionally, $W$ refers to a weighting matrix of shape $h \times w$, characterized by values decreasing from $1$ at its center to $0$ towards the edges.
These quantities are combined element-wise through the Hadamard product $\odot$.

The reward function encourages the robot to maneuver its camera closer to the target and ensure the target remains centered within the frame. For consistency across experiments, we set $c = 800$ and bound the rewards between $0$ and $4$.

Each episode spans $150$ timesteps, with each timestep lasting $40$ ms, totaling $6$ seconds. 
The agent sends an action every timestep, which the SenseAct system repeats five times every $8$ ms.

\end{document}